\documentclass{article}

\PassOptionsToPackage{numbers, compress}{natbib}

\usepackage[preprint]{neurips_2025}

\usepackage[utf8]{inputenc} 
\usepackage[T1]{fontenc}    
\usepackage{hyperref}       
\usepackage{url}            
\usepackage{booktabs}       
\usepackage{amsfonts}       
\usepackage{nicefrac}       
\usepackage{microtype}      
\usepackage{xcolor}         
\usepackage{xpatch}
\usepackage{graphicx}
\usepackage{multirow}
\usepackage[table]{xcolor}
\usepackage{pifont}
\usepackage{makecell}
\usepackage{amsmath} 
\usepackage{subcaption}
\usepackage[most]{tcolorbox}

\definecolor{lightyellow}{RGB}{255, 244, 230}
\definecolor{mycyan}{HTML}{FFF0F6}
\definecolor{mycyan1}{HTML}{F783AC}
\hypersetup{
    colorlinks=true,
    citecolor=teal,
    linkcolor=red,
    filecolor=magenta,
    urlcolor=magenta}

\makeatletter
\xapptocmd{\NAT@bibsetnum}{\setlength{\leftmargin}{0pt}\setlength{\itemindent}{\labelwidth}\addtolength{\itemindent}{\labelsep}}{}{}
\makeatother

\title{From Recognition to Understanding: Unlocking Cognitive Time Series Reasoning with LLMs}

\author{%
  Xin Qiu$^{1,2}$, Junlong Tong$^{1}$, Yao Zhang$^{1}$, Yunpu Ma$^{3}$, Wei Zhang$^{1}$, 
Xiaoyu Shen$^{1}$\thanks{Corresponding author.} \\
  $^1$Institute of Digital Twin, Eastern Institute of Technology, Ningbo\\
   $^2$Zhejiang University\quad
   $^3$Munich Center for Machine Learning, LMU\\
  \texttt{qiuxinzju@zju.edu.cn}\quad\texttt{xyshen@eitech.edu.cn} \\
}

\begin{document}

\maketitle

\begin{abstract}
Time series analysis has recently been coupled with Large Language Models (LLMs) to leverage their reasoning and world knowledge capabilities, yet gains remain limited. We attribute this to a fundamental mismatch between existing task formulations and LLM strengths: most settings reduce time series understanding to curve-fitting systems, focusing on low-level prediction while ignoring the semantic, contextual, and reasoning-intensive nature of real-world temporal decision-making.
To address these limitations, we introduce \textbf{\textit{TSCognition}}, a multimodal benchmark for multi-dimensional time series reasoning. It collects real-world time series and textual information from \textbf{\textit{15}} public sources and constructs \textbf{\textit{$\sim$41K}} QA samples around five cognitive reasoning tasks: Decoding, Grounding, Inferring, Extrapolating, and Acting. Building on this, we further propose \textbf{\textit{TSAlign}}, a unified framework that encodes time series into compact patch-level representations and aligns them with semantic directions in the LLM embedding space via gated residual injection and multivariate fusion.
Experiments show that TSAlign \textbf{\textit{outperforms}} existing LLM, VLM, and time series QA baselines on our TSCognition and the publicly available TimerBed while substantially reduce computational cost. 
Code is available at: https://github.com/EIT-NLP/CognitiveTSR
\end{abstract}
\section{Introduction}

Time series analysis models temporal dependencies in sequential data to uncover latent patterns and forecast future behavior, playing a critical role across diverse real-world domains~\cite{brown2020language,TONG2022116049,qiu2025few,jiang2025multi}. Recognizing the profound success of Large Language Models (LLMs), researchers have increasingly sought to integrate them into time series analysis, hoping to leverage their vast pre-trained knowledge to enhance scenario understanding~\cite{qiu2026rethinking,kong2025time,wang2025itformer}.

However, despite efforts across tasks such as forecasting, classification, and anomaly detection~\cite{liu2025timecma,jiang2025fstllm,jia2026m3time}, introducing LLMs to time series domains has yielded limited improvements. Emerging evidence suggests this bottleneck stems from a fundamental mismatch between current task formulations and the intrinsic capabilities of LLMs~\cite{zheng2026lifting,xie2025chatts,tan2024language}. Specifically, traditional tasks mainly focus on shallow pattern mining, such as predicting future points, imputing missing values, or detecting anomalies. In practice, however, real-world time series are inherently stochastic and heavily influenced by unobserved external factors, making precise trajectory prediction fundamentally ill-posed. Ultimately, these setups reduce LLMs to sophisticated curve-fitters, while their true strengths of extensive world knowledge, contextual understanding, and complex reasoning remain completely unutilized~\cite{liu2024time,wu2025scits}.

To better align time series analysis with LLM capabilities, recent work advocates a shift from isolated prediction toward language-conditioned temporal reasoning~\cite{yu2026tsrbench,chen2025mtbench,guan2026timeomni,wu2026timeart}. This paradigm integrates temporal signals with natural language queries and external context, requiring models to perform evidence extraction, knowledge grounding, and logical inference over time-evolving systems.

However, this transition is fundamentally constrained by the lack of suitable data. Most existing datasets remain designed for forecasting or anomaly detection~\cite{guan2026timeomni,wu2025scits,yu2026tsrbench}, while early time series QA benchmarks such as TimeOmni-1~\cite{guan2025timeomni} and Time-MQA~\cite{kong2025time}  short univariate time series, shallow questions, and narrow domains, lacking semantic depth and compositional reasoning, as shown in Fig.~\ref{fig:limitations}. This limits their ability to evaluate higher-order  cognition.

To address this gap, we introduce \textbf{\textit{TSCognition}}, a multimodal dataset designed for multi-dimensional time series reasoning. Constructed from 15 public sources, it pairs real-world time series with rich contextual text and contains 41,086 samples spanning five reasoning tasks. Compared to prior datasets, TSCognition evaluates finer-grained temporal cognition, covering pattern decoding, context grounding, relational inference, future extrapolation, and decision-oriented action.

Beyond the data bottleneck, existing methodologies for integrating time series with LLMs face severe representational limitations. Existing methods usually convert time series into text, images, or projected features, which may fragment temporal structures, distort numerical dynamics, and lack language-space alignment~\cite{nie2022time,nagrath2026patch,liu2025picture}.
To overcome this representational flaw and efficiently bridge continuous time series representations with the discrete language reasoning space, we propose \textbf{\textit{TSAlign}}. TSAlign is motivated by a simple intuition: time series tokens should preserve temporal dynamics while remaining understandable to the LLM’s pretrained language space. It learns compact patch-level representations and explicitly aligns them with principal semantic directions in the LLM embedding space. A gated residual mechanism then injects semantic guidance without overwriting the original temporal information, reducing the information loss caused by direct feature replacement. For multivariate inputs, TSAlign further adopts gated fusion to adaptively preserve multiple informative variables. The fused temporal tokens are finally fed into the LLM together with textual inputs, such as questions and candidate answers, for reasoning.

Extensive experiments demonstrate that TSAlign achieves consistent advantages in both QA reasoning and traditional pattern analysis. On TSCognition, TSAlign outperforms existing LLM, VLM, and time series QA baselines while drastically reducing token overhead. It uses \textbf{\textit{3.0$\times$}} fewer tokens than vision-based inputs and \textbf{\textit{16.8$\times$}} fewer than textual inputs. Furthermore, on the established TimerBed benchmark, TSAlign achieves state-of-the-art performance in pattern analysis, proving that \emph{when properly aligned and tasked, LLMs significantly outperform traditional time series models}. 

In summary, our main contributions are as follows:
\begin{enumerate}
\item \emph{A Cognitively Structured Dataset:} We curate \textbf{\textit{TSCognition}}, integrating real-world time series with contextual text to evaluate LLMs across five hierarchical reasoning tasks, moving beyond shallow pattern mining.
\item \emph{A Unified Alignment Framework:} We propose \textbf{\textit{TSAlign}}, which bridges time series and language reasoning spaces through semantic alignment, gated residual preservation, and multi-dimensional fusion, avoiding the pitfalls of numerical serialization.
\item \emph{State-of-the-Art Performance \& Efficiency:} We demonstrate that TSAlign delivers superior reasoning accuracy and highly efficient token utilization on both the novel TSCognition dataset and the public TimerBed benchmark.
\end{enumerate}
\section{Related Work}
\textbf{Multi-Modal QA.}\quad
Recent advances in LLMs have substantially improved natural language understanding and further promoted multimodal learning~\cite{xu2025llava,zhangllava,liu2026vica}. In particular, multimodal QA has been largely driven by vision-language benchmarks, where successful answering requires models to integrate visual evidence~\cite{li2023time,tarasiou2023vits,he2026harnessing,xu2026can}. However, simply applying a vision--language paradigm is not an ideal way to handle temporal problems~\cite{guan2026timeomni,ni2025harnessing}. Although this strategy can leverage the capabilities of existing multimodal models, temporal dependencies are often distorted during the imaging process~\cite{prithyani2024feasibility,zhong2025time}. Another line of work attempts to directly discretize or textualize time series and feed them into LLMs together with natural language, but this also suffers from clear limitations, due to the inherent gap between continuous numerical signals and discrete linguistic symbols~\cite{guan2025timeomni,kong2025time}. Therefore, how to support multivariate, multi-step time series QA remains a challenging problem.

\textbf{Language Models for Time Series QA.}\quad 
Existing attempts to connect language with time series data have mostly used text as a supplementary signal for enhancing conventional tasks, such as forecasting and classification, rather than treating language as a core component of temporal reasoning~\cite{ICLR2024_680b2a81,liu2025calf,meunier2025crisists}. In many cases, these methods are built upon handcrafted prompts or overly simplified experimental settings, which weakens their applicability to realistic scenarios.
Existing language-based time series studies mostly focus on surface descriptions or basic pattern interpretation~\cite{guan2025timeomni,wu2025scits,chen2025mtbench,yu2026tsrbench}. Moreover, textualizing time series for LLMs is computationally expensive and poorly suited to real-time, multivariate, or long-horizon analysis~\cite{cheng2026can,luo2025time}. More fundamentally, such formulations overlook the intrinsic modality gap between time series and natural language, thereby limiting the model’s ability to capture the underlying temporal structures effectively~\cite{schumacher2026prompting}.

\section{Background}
\begin{figure}[t]
    \centering
    \includegraphics[width=1\linewidth]{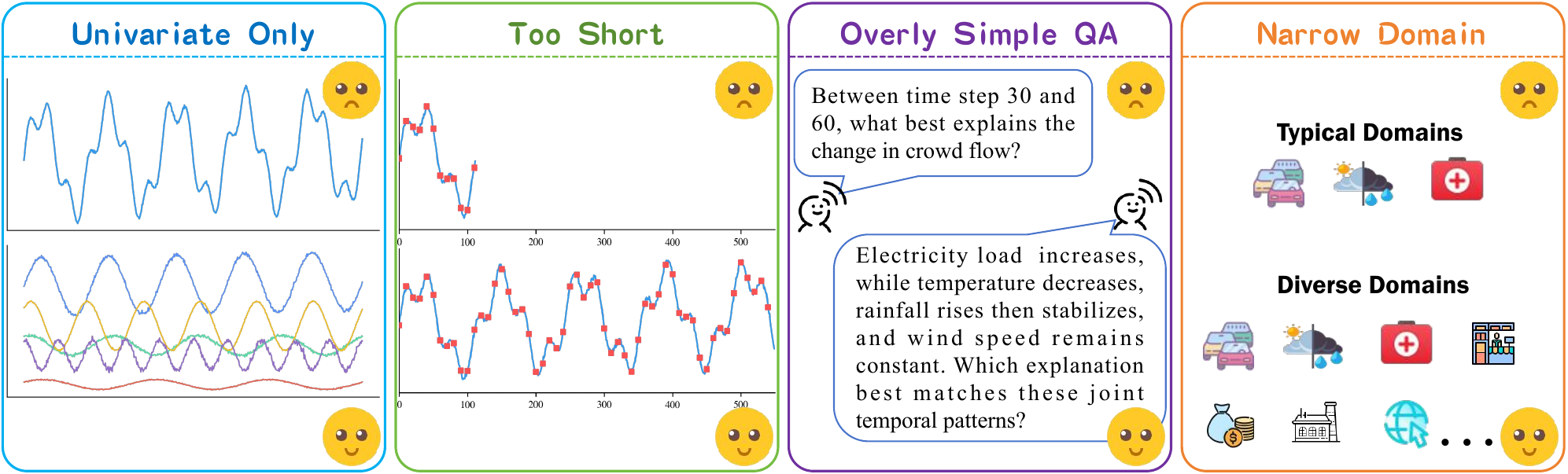}
    \caption{Limitations of existing time series QA datasets (upper) vs. our TSCognition (bottom). Our TSCognition introduces compositional reasoning over realistic real-world scenarios, better testing whether models truly understand over temporal patterns rather than relying on superficial cues.}
    \label{fig:limitations}
\end{figure}
\subsection{Problem Formulation}
We first formalize the LLM-based time series understanding problem considered in this work. Given a multivariate time series
$
X = \{x_1, x_2, \dots, x_T\},  x_t \in \mathbb{R}^d,
$
where $T$ denotes the sequence length and $d$ denotes the variable dimension. Rather than treating time series merely as numerical sequences, LLM-based time series understanding aims to capture temporal structures and connect them with semantic concepts, task conditions, and decision objectives.
In this context, evaluations are instantiated in two forms: \textbf{\textit{QA reasoning tasks}} and \textbf{\textit{pattern analysis tasks}}. 

For \textbf{QA reasoning tasks}, let the natural language query be denoted as
$
q \in \mathcal{Q},
$
where $q$ may contain task instructions, contextual information, background descriptions, or reasoning objectives. The model is required to generate an answer
$
a \in \mathcal{A}
$
conditioned on both the time series and the query. This can be formulated as
$
f_{\theta}: (X, q) \rightarrow a,
$
or equivalently as the conditional distribution
$
p_{\theta}(a \mid X, q).
$
To characterize this process, we introduce a latent reasoning state $z$, which represents the intermediate understanding induced jointly by temporal evidence and query semantics:
$
p_{\theta}(a \mid X, q)
=
\int p_{\theta}(a \mid z, q)\,
p_{\theta}(z \mid X, q)\, dz.
\label{eq:qa_reasoning}
$
$p_{\theta}(z \mid X, q)$ models the construction of a latent reasoning state from the time series and the query, while $p_{\theta}(a \mid z, q)$ generates the final answer conditioned on this state. The  training objective is:
\begin{equation}
\min_{\theta} \ 
\mathbb{E}_{(X, q, a) \sim \mathcal{D}_{\mathrm{QA}}}
\left[
-\log p_{\theta}(a \mid X, q)
\right].
\label{eq:qa_obj}
\end{equation}

For \textbf{pattern analysis tasks}, the model is expected to directly identify and analyze temporal patterns from the raw time series, without relying on an explicit natural language query. Let
$
y \in \mathcal{Y}
$
denote the pattern label or analysis target associated with $X$. The task can be formulated as
$
f_{\theta}: X \rightarrow y,
$
or probabilistically as
$
p_{\theta}(y \mid X).
$
Similarly, we introduce a latent pattern representation $z$ to denote the structured temporal information extracted from the input sequence:
$
p_{\theta}(y \mid X)
=
\int p_{\theta}(y \mid z)\,
p_{\theta}(z \mid X)\, dz.
\label{eq:pa_reasoning}
$
$p_{\theta}(z \mid X)$ extracts task-relevant temporal structures from the raw time series, while $p_{\theta}(y \mid z)$ maps the inferred pattern representation to the final analysis target. The corresponding training objective is:
\begin{equation}
\min_{\theta} \ 
\mathbb{E}_{(X, y) \sim \mathcal{D}_{\mathrm{PA}}}
\left[
-\log p_{\theta}(y \mid X)
\right].
\label{eq:pa_obj}
\end{equation}

Therefore, pattern analysis evaluates the extraction of discriminative temporal structures~\cite{yue2022ts2vec,qiao2026s}, while QA reasoning evaluates their alignment with language semantics and task objectives~\cite{jing2026tsaqa,chen2025mtbench}.

\subsection{Revisiting the Limitations of TS Reasoning Dataset}
Despite recent efforts to incorporate natural language into time series analysis, existing time series reasoning settings remain fundamentally limited~\cite{kong2025time,wang2025itformer,chen2025mtbench,guan2025timeomni}. We analyze these issues from a structural perspective, as shown in Fig.~\ref{fig:limitations}.
\textbf{\textit{(1) Most existing approaches are primarily built upon univariate time series, where each sample contains only a single variable evolving over time.}} This formulation overlooks the multi-variable interactions and auxiliary information that are pervasive in real-world scenarios, such as inter-variable dependencies, external environmental factors, and cross-modal signals.
\textbf{\textit{(2) The time series used in existing datasets are typically short, often limited to around 50–125 time steps.}} Under such settings, many tasks can be solved through local pattern matching or simple statistical cues, without requiring a genuine understanding of global temporal structures.
\textbf{\textit{(3) The design of existing time series QA tasks is often overly simplistic and surface-level.}} Questions typically correspond to direct queries or single-step inference, allowing models to rely on shallow pattern recognition rather than engaging in multi-step reasoning. As a result, strong performance on these benchmarks does not necessarily indicate true reasoning capability.
\textbf{\textit{(4) The range of application domains covered by existing benchmarks remains relatively narrow.}} The data often exhibit fixed statistical properties and task formats, lacking the diversity of cross-domain and multi-scenario settings. 
\textbf{In summary}, existing time series reasoning tasks remain limited in structure, length,  complexity, and domain coverage.
\begin{figure}[t]
    \centering
    \includegraphics[width=1\linewidth]{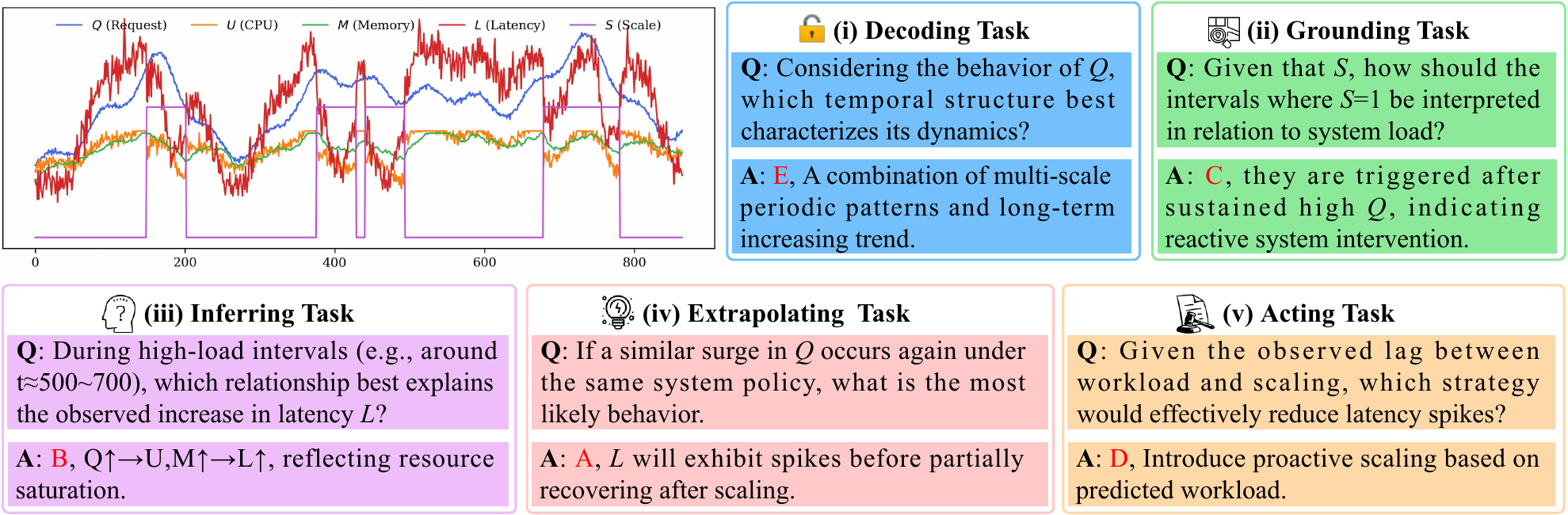}
    \caption{Using a cloud system workload scenario as an example, we construct five hierarchical tasks based on multivariate time series ($Q,U,M,L,S$): Decoding, Grounding, Inferring, Extrapolating, and Acting, to characterize the full reasoning process from pattern understanding to decision-making.}
    \label{fig:reasonTS}
    \vspace{-0.3cm}
\end{figure}

\subsection{TSCognition: A Multimodal Dataset for Multi-Dimensional TS Reasoning}
\label{TSCognition}
To address these limitations, we introduce \textbf{\textit{TSCognition}}. It enables more realistic and rigorous evaluation of model generalization and reasoning capabilities. \textit{To the best of our knowledge, this is the first multimodal dataset for time series that unifies multi-dimensional data, long-horizon dynamics, and cognitively structured reasoning tasks.} 

\textbf{Domain Coverage.}\quad
We collect raw data from 15 publicly available sources to construct a multimodal dataset. Each sample consists of time series signals paired with corresponding textual information, such as event descriptions, contextual backgrounds, or task instructions.
TSCognition covers 8 domains, including industry, web, finance, climate, transportation, healthcare, retail, and energy. These domains exhibit substantial diversity in data distributions, temporal dynamics, and task requirements.
Overall, its multi-source, multi-domain design ensures broad coverage of distributions, dynamics, and semantics, enabling robust evaluation of generalization.

\textbf{Task Taxonomy.}\quad
To systematically characterize the cognitive structure of time series reasoning, we decompose it into five hierarchical tasks: \textbf{\textit{Decoding}}, \textbf{\textit{Grounding}}, \textbf{\textit{Inferring}}, \textbf{\textit{Extrapolating}}, and \textbf{\textit{Acting}}, as shown in Fig.~\ref{fig:reasonTS}.

\textbf{\textit{(i) Decoding task}} extracts fundamental temporal patterns from raw series, including trends, periodicity, and local structures, providing the basis for subsequent reasoning. \textbf{\textit{(ii) Grounding task}} aligns time series with contextual information or task descriptions, enabling context-aware interpretation rather than isolated analysis. \textbf{\textit{(iii) Inferring task}} performs relational modeling and multi-step reasoning, capturing dependencies, causal cues, and conditional relationships. \textbf{\textit{(iv) Extrapolating task}} extends inferred structures to future dynamics under uncertainty, emphasizing structured reasoning over simple numerical fitting. \textbf{\textit{(v) Acting task}} integrates preceding results to produce goal-oriented decisions, supporting adaptive actions in complex environments.

\textbf{Data Statistics.}\quad
TSCognition contains a total of 41,086 QA pairs, with a train–evaluation split ratio of 0.8:0.2. Specifically, 32,869 samples are used for training, while 8,217 samples are reserved for evaluation, including 6,163 in-distribution (ID) samples and 2,054 out-of-distribution (OOD) samples.
In terms of data complexity, each question is associated with multivariate time series whose dimensionality ranges from 3 to 6 variables. The length of each individual time series dimension varies from 437 to 1059 time steps, with an average length of 785. In addition, each question is formulated as a multiple-choice problem with 4 or 5 options, among which only one is correct.

\textbf{Pipeline.}\quad
In addition, we design a pipeline, which consists of four stages: \textbf{\textit{raw data collection}}, \textbf{\textit{data cleaning}}, \textbf{\textit{QA construction}}, and \textbf{\textit{quality control}}. During data collection, we prioritize time series aligned with corresponding textual information to ensure cross-modal dependence. We then apply unified cleaning by removing samples with missing values, timestamp errors, ordering issues, modality mismatches, duplicates, invalid fields, and noisy descriptions, before pairing the remaining data into structured time series--text units. Based on the five task categories, we construct QA samples with plausible distractors derived from local patterns, semantically similar descriptions, or common misinterpretations. Finally, each sample undergoes multi-round human verification, where disagreements are resolved by an additional reviewer and any sample failing the quality criteria is discarded, ensuring reliability and reducing bias. 

\section{TSAlign: A Framework for Bridging TS and Language Reasoning}
\begin{figure}[t]
    \centering
    \includegraphics[width=1\linewidth]{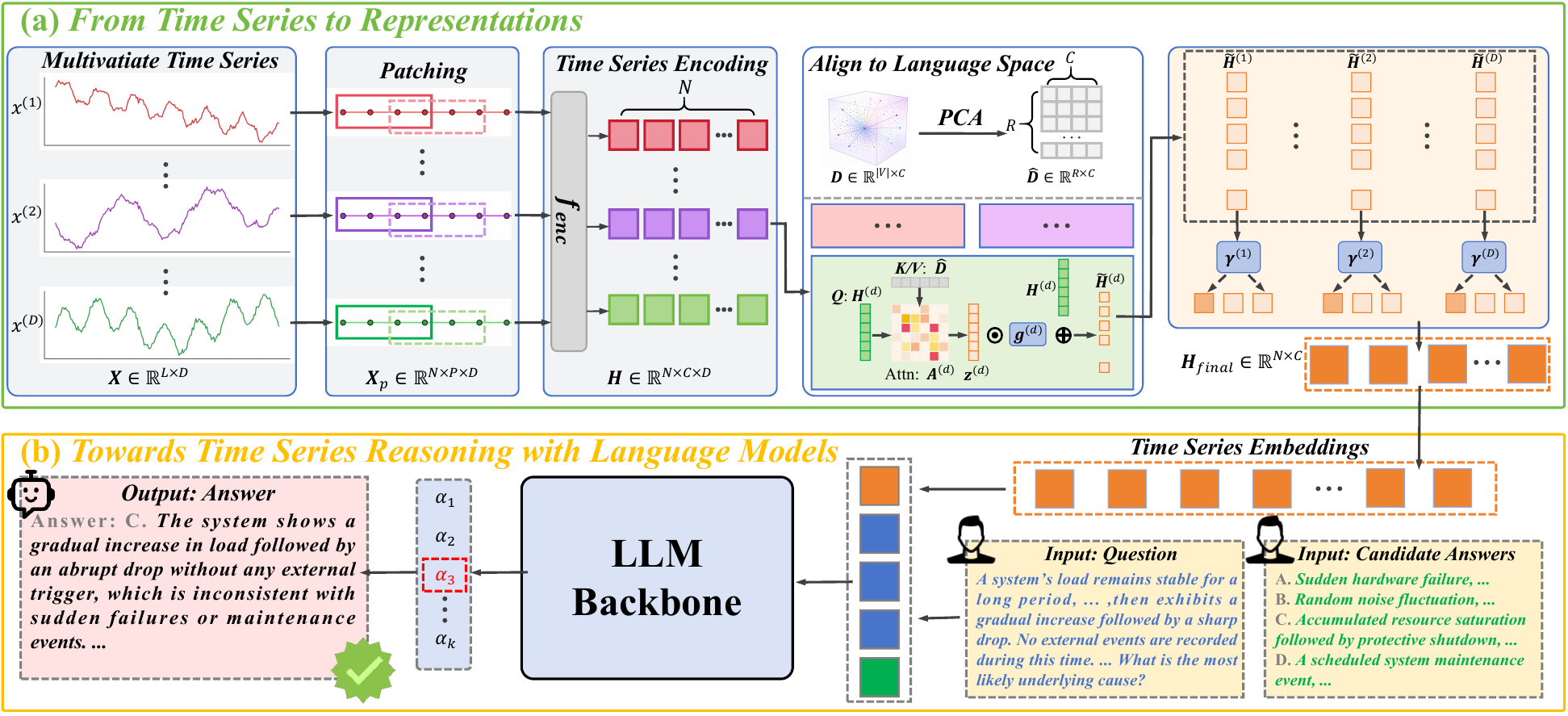}
    \caption{Overview of TSAlign. TSAlign encodes multivariate time series into patch-level representations, aligns them with the language space, and fuses variables into unified tokens for reasoning.}
    \label{fig:main_figure}
\end{figure}
Fig.~\ref{fig:main_figure} provides an overview of TSAlign, which first derives language-compatible time series representations and then integrates them with textual inputs for LLM-based reasoning.
\subsection{From Time Series to Representations}
\label{From Time Series to Representations}
Given a multivariate time series $\mathbf{X} \in \mathbb{R}^{L \times D}$, where $L$ denotes the number of time steps and $D$ the number of variables, the objective is to derive a unified representation for downstream reasoning.

\textbf{Patch-based Temporal Encoding.}\quad
Modeling long time series is challenging due to varying numbers of time steps and complex temporal dependencies. To address this, we decompose each variable $\mathbf{x}^{(d)} \in \mathbb{R}^{L}$ into a sequence of overlapping patches with length $P$ and stride $S$~\cite{nie2022time}, yielding
$
\mathbf{x}^{(d)}_p \in \mathbb{R}^{N \times P}, 
N = \left\lfloor \frac{L - P}{S} \right\rfloor + 1.
$
Each patch is then encoded by a time series encoder $f_{\mathrm{enc}}(\cdot)$:
$
\mathbf{H}^{(d)} = f_{\mathrm{enc}}(\mathbf{x}^{(d)}_p), 
\mathbf{H}^{(d)} \in \mathbb{R}^{N \times C},
$
where $C$ denotes the embedding dimension. This results in a sequence of patch-level embeddings for each variable.

\textbf{Alignment to Language Space.}\quad
Given the discrepancy between time series and language embedding spaces, we align the former to the semantic structure of the pre-trained language embedding space to improve representation compatibility. Let $\mathbf{D} \in \mathbb{R}^{|\mathcal{V}| \times C}$ denote the language embedding matrix, where $|\mathcal{V}|$ is the vocabulary size. Due to its large scale, we apply principal component analysis (PCA) to obtain a compact set of principal semantic directions:
$
\hat{\mathbf{D}} = \mathrm{PCA}(\mathbf{D}),  
\hat{\mathbf{D}} \in \mathbb{R}^{R \times C},  R \ll |\mathcal{V}|.
$
We then perform attention-based alignment by treating the time series embeddings as queries and the principal language embeddings as keys and values:
\begin{equation}
\mathbf{A}^{(d)} = \mathrm{softmax}\left({\mathbf{H}^{(d)} \hat{\mathbf{D}}^\top}/{\sqrt{C}}\right),\quad
\mathbf{Z}^{(d)} = \mathbf{A}^{(d)} \hat{\mathbf{D}}.
\end{equation}

To preserve patch-level temporal information while incorporating semantic guidance, we formulate alignment as a gated residual update. The original time series embedding is kept as the backbone representation, and the language-embedding-induced semantic representation is added as a controlled correction whose strength is determined by a learnable gate:
\begin{equation}
\mathbf{g}^{(d)} = \sigma(\mathbf{H}^{(d)} \mathbf{w}_g), \quad
\tilde{\mathbf{H}}^{(d)} = \mathbf{H}^{(d)} + \mathbf{g}^{(d)} \odot \mathbf{Z}^{(d)}.
\end{equation}
The resulting $\tilde{\mathbf{H}}^{(d)} \in \mathbb{R}^{N \times C}$ preserves sequence length while aligning with the language embedding space, enabling segment-wise semantic refinement without information loss.

\textbf{Multi-dimensional Fusion.}\quad
Following alignment, each variable is represented as a sequence of embeddings $\tilde{\mathbf{H}}^{(d)}$. Since the number of variables $D$ may vary across samples, a unified representation must be constructed.
To this end, we modulate the contribution of each dimension through a gating mechanism. Specifically, we first compute a summary representation for each dimension:
$
\bar{\mathbf{h}}^{(d)} = \frac{1}{N} \sum_{i=1}^{N} \tilde{\mathbf{H}}^{(d)}_i,
$
and assign an importance weight to the $d$-th dimension:
$
g^{(d)} = \sigma(w_g^\top \bar{\mathbf{h}}^{(d)} + b_g).
$
The final representation is obtained by weighted fusion across dimensions at each patch position:
\begin{equation}
\mathbf{H}_{\mathrm{final}, i}
=
\frac{1}{\sum_{d=1}^{D} g^{(d)}}
\sum_{d=1}^{D} g^{(d)} \tilde{\mathbf{H}}^{(d)}_i,
\quad i=1,\dots,N.
\end{equation}
Thus, the fused representation is
$
\mathbf{H}_{\mathrm{final}} \in \mathbb{R}^{N \times C}.
$
This formulation preserves patch-level structure while enabling flexible multi-dimensional integration. Independent gating avoids competitive normalization, allowing multiple informative dimensions to contribute. The resulting $\mathbf{H}_{\text{final}}$ serves as a consistent representation for downstream reasoning.

\subsection{Towards Time Series Reasoning with Language Models}
Given the unified time series representation $\mathbf{H}_{\text{final}} \in \mathbb{R}^{N \times C}$ obtained in Sec.~\ref{From Time Series to Representations}, we combine it with textual inputs to perform reasoning via a LLM. Each sample consists of a question $q$ and a set of candidate answers $\mathcal{A}=\{a_1,\dots,a_K\}$, where the question may further include task instructions and contextual descriptions.
To bridge numerical time series and language tokens, we first project the time series representation into the embedding space of the LLM:
$
\tilde{\mathbf{z}} = W_p \mathbf{z},
$
where $W_p \in \mathbb{R}^{d \times C}$ and $d$ denotes the hidden size of the LLM. The projected vector $\tilde{\mathbf{z}} \in \mathbb{R}^{d}$ is treated as a special temporal token that summarizes the global dynamics of the input series.
We then concatenate this temporal token with the token embeddings of the textual question and candidate answers to form a unified input sequence:
$
\mathbf{E}_k = \mathcal{F}\big(\tilde{\mathbf{z}},\ \mathrm{Emb}(q),\ \mathrm{Emb}(a_k)\big),
$
where $\mathrm{Emb}(\cdot)$ denotes the token embedding function of the LLM, and $\mathbf{E}_k$ is the resulting multimodal representation for the candidate answer $a_k$.

The resulting embedding sequence is fed into the LLM, which induces a scoring function over candidate answers:
$
s_k = f_{\theta}(\mathbf{E}_k),
$
where $f_{\theta}$ denotes the LLM parameterized by $\theta$. The final prediction is given by:
$
\hat{y} = \arg\max_k s_k.
$
This formulation enables the model to condition explicitly on temporal representations rather than relying on textualized or visualized surrogates of time series, allowing the LLM to jointly model temporal structures and linguistic semantics for more effective reasoning. 

We train the model in a supervised manner using the multiple-choice question answering formulation. Given a question $q$ and a set of candidate answers $\mathcal{A} = \{a_1, \dots, a_K\}$, the model produces a score $s_k$ for each candidate $a_k$ based on the joint reasoning over time series and textual inputs.
During training, we adopt an end-to-end optimization strategy, jointly updating the time series encoder, projection layer, and the LLM backbone.

\section{Experiment}
\begin{table*}[t]
\centering
\caption{Accuracy (\%) on five QA reasoning task categories under ID and OOD settings.}
\resizebox{1\textwidth}{!}{
\renewcommand{\arraystretch}{1.0}
\begin{tabular}{c|c|cc|cc|cc|cc|cc}
\toprule
\midrule
\multirow{2}{*}{\textbf{Methods}} & \multirow{2}{*}{\textbf{Types}} 
& \multicolumn{2}{c|}{\textbf{Decoding}} 
& \multicolumn{2}{c|}{\textbf{Grounding}} 
& \multicolumn{2}{c|}{\textbf{Inferring}} 
& \multicolumn{2}{c|}{\textbf{Extrapolating}} 
& \multicolumn{2}{c}{\textbf{Acting}} \\
& 
& \textbf{ID} & \textbf{OOD} 
& \textbf{ID} & \textbf{OOD}
& \textbf{ID} & \textbf{OOD}
& \textbf{ID} & \textbf{OOD}
& \textbf{ID} & \textbf{OOD} \\
\midrule
\textbf{\textit{Random Guessing}} & -- &22.1 &23.5&22.4&21.9&24.2&21.8&23.5&22.7&24.1&23.9 \\
\midrule
\multicolumn{12}{c}{\textbf{Zero-shot Test}} \\
\midrule
\rowcolor{lightyellow}\textbf{\textit{GPT-5.1}} & Proprietary API & 71.4 & 68.5 & \textcolor{blue}{74.6} & 70.1 & \textcolor{blue}{68.3} & 72.5 & 57.4 & 63.9 & 71.5 & 77.0 \\

\textbf{\textit{Qwen2.5-Instruct-32B}} & TS-Text & 43.2 & 43.2 & 38.3 & 39.5 & 40.7 & 36.8 & 35.2 & 33.6 & 37.2 & 34.9 \\
\rowcolor{lightyellow}\textbf{\textit{Qwen2.5-Instruct-7B}} & TS-Text & 34.6 & 37.1 & 29.4 & 32.7 & 36.5 & 33.3 & 28.3 & 30.2 & 32.6 & 29.7 \\
\textbf{\textit{Qwen2.5-VL-7B}} & Vision-Text & 37.5 & 41.2 & 34.1 & 34.7 & 30.7 & 28.4 & 33.0 & 34.8 & 26.9 & 25.3 \\
\midrule
\multicolumn{12}{c}{\textbf{Full-shot Test}} \\
\midrule

\rowcolor{lightyellow}\textbf{\textit{Qwen2.5-Instruct-7B}} & TS-Text & 51.7 & 62.3 & 44.0 & 48.6 & 45.2 & 41.7 & 39.8 & 36.5 & 41.5 & 45.6 \\
\textbf{\textit{Qwen2.5-VL-7B}} & Vision-Text & 48.5 & 52.6 & 40.2 & 46.7 & 39.5 & 35.4 & 38.1 & 41.5 & 38.2 & 32.5 \\

\midrule

\rowcolor{lightyellow}\textbf{\textit{ITFormer}} & TS-Text & 42.4 & 46.8 & 47.4 & 50.6 & 42.1 & 43.6 & 46.5 & 42.7 & 40.8 & 42.3 \\
\textbf{\textit{Time-MQA}} & TS-Text & 45.5 & 44.1 & 42.5 & 41.4 & 36.5 & 38.4 & 34.8 & 31.4 & 36.8 & 32.7 \\
\rowcolor{lightyellow}\textbf{\textit{Time-LLM}} & TS-Text & 36.0 & 38.2 & 32.5 & 36.1 & 34.1 & 32.8 & 35.4 & 39.3 & 31.6 & 33.4 \\
\textbf{\textit{ChatTS}} & TS-Text & 39.1 & 43.5 & 36.8 & 42.3 & 37.4 & 30.8 & 32.0 & 36.8 & 28.5 & 30.6 \\
\rowcolor{lightyellow}\textbf{\textit{GPT4TS}} & TS-Text & 34.8 & 37.9 & 33.2 & 38.4 & 32.4 & 29.7 & 31.2 & 33.5 & 35.6 & 38.1 \\

\midrule

\textbf{\textit{TSAlign-3B}} & Ours & \textcolor{blue}{74.9} & \textcolor{blue}{76.5} & 72.7 & \textcolor{blue}{76.6} & 65.8 & \textcolor{blue}{74.3} & \textcolor{blue}{74.5} & \textcolor{blue}{81.3} & \textcolor{blue}{82.4} & \textcolor{blue}{86.0} \\
\rowcolor{lightyellow}\textbf{\textit{TSAlign-7B}} & Ours & \textcolor{red}{83.1} & \textcolor{red}{87.6} & \textcolor{red}{85.4} & \textcolor{red}{87.4} & \textcolor{red}{91.7} & \textcolor{red}{88.7} & \textcolor{red}{77.9} & \textcolor{red}{84.6} & \textcolor{red}{90.4} & \textcolor{red}{95.3} \\
\midrule
\bottomrule
\end{tabular}
}
\label{tab:main_results}
\vspace{-0.5cm}
\end{table*}

\subsection{Comparison on QA Reasoning Tasks}
\textbf{Baselines \& Evaluation}.\quad 
We compare our method with several representative baselines, covering general-purpose LLMs, VLMs, and existing time series QA models. 
For general LLMs and VLMs, we consider \textbf{\textit{GPT-5.1}}, \textbf{\textit{Qwen2.5-Instruct}}~\cite{qwen2025qwen25technicalreport}, and \textbf{\textit{Qwen2.5-VL}}~\cite{qwen2025qwen25technicalreport} as representative models. For these models, we evaluate two typical input paradigms: (1) representing time series as text and combining them with background information (TS-Text), and (2) converting time series into line charts and providing them together with background information (Vision-Text). 
For time series QA methods, we include several recent approaches, including \textbf{\textit{ITFormer}}~\cite{wang2025itformer}, \textbf{\textit{Time-MQA}}~\cite{kong2025time}, \textbf{\textit{Time-LLM}}~\cite{ICLR2024_680b2a81}, \textbf{\textit{ChatTS}}~\cite{xie2025chatts}, and \textbf{\textit{GPT4TS}}~\cite{zhou2023one}. For fair comparison, all methods are implemented using Qwen2.5-Instruct-7B as the backbone LLM.
In our \textbf{\textit{TSAlign}} family, the 3B \& 7B variants use Qwen2.5-Instruct-3B \& 7B as the backbone LLMs. 
We evaluate TSCognition under both zero-shot and full-shot settings, where models are either directly tested without task-specific training or trained before evaluation.

\textbf{Main Results}.\quad 
As shown in Tab.~\ref{tab:main_results}, the proposed TSAlign outperforms existing methods across all five reasoning task categories. Specifically, under the full-shot setting, TSAlign-7B achieves the best performance, with particularly notable gains on higher-level reasoning tasks such as Inferring and Acting. Under the zero-shot setting, TSAlign-3B already surpasses it on the majority of tasks. Quantitatively, TSAlign-7B outperforms TS-Text baselines with the same backbone by \textbf{42.3\%}, \textbf{43.0\%}, \textbf{47.3\%}, \textbf{46.7\%}, and \textbf{48.1\%} across the five tasks, while TSAlign-3B still achieves gains of \textbf{34.1\%}, \textbf{34.8\%}, \textbf{39.1\%}, \textbf{38.5\%}, and \textbf{39.9\%}. Moreover, TSAlign-7B surpasses the proprietary GPT-5.1 by an average margin of \textbf{17.7\%}.
This observation suggests that, our model can sufficiently leverage the structured knowledge encoded in LLMs to perform time series reasoning.

\begin{figure}[t]
    \centering
    \vspace{-0.35cm}
    \includegraphics[width=1\linewidth]{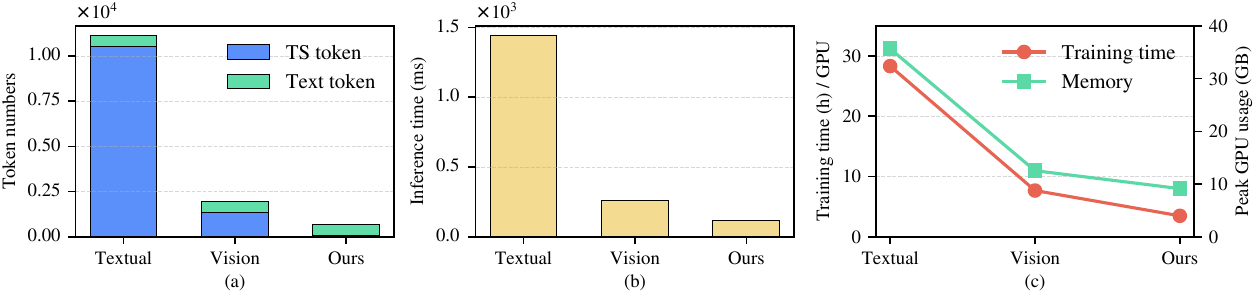}
    \caption{Efficiency analysis. (a) Token composition (TS token vs. Text token); (b) Average inference time per sample; (c) Training time normalized to a single A800 GPU (left) and peak memory (right).}
    \label{fig:three_plots_nips}
    \vspace{-0.25cm}
\end{figure}

\textbf{Efficiency Analysis}.\quad 
As shown in Fig.~\ref{fig:three_plots_nips}, we compare the efficiency of TSAlign-7B with two common paradigms that convert time series into textual or vision inputs for QA reasoning under a unified experimental setup. Specifically, all methods are evaluated with the same hyperparameter configuration, using Qwen2.5-Instruct-7B and Qwen2.5-VL-7B as the base LLMs for textual and visual paradigms.
TSAlign significantly reduces token complexity, which is \textbf{2.96$\times$} fewer than vision-based inputs and \textbf{16.79$\times$} fewer than textual inputs. This leads to consistent computational gains: TSAlign achieves \textbf{2.23$\times$} and \textbf{12.31$\times$} faster inference, and \textbf{2.21$\times$} and \textbf{8.14$\times$} faster training compared to vision and textual methods, respectively. 
Moreover, TSAlign requires the lowest memory, reducing peak GPU usage, which is \textbf{1.37$\times$} and \textbf{3.90$\times$} lower than vision and textual baselines.

\subsection{Comparison on Pattern Analysis Tasks}
\textbf{Baselines \& Evaluation}.\quad
To evaluate performance on time series pattern analysis tasks, we compare \textbf{\textit{TSAlign}} with representative methods from three paradigms.
(1) General LLMs, including the proprietary \textbf{\textit{GPT-5.1}} and the open-source \textbf{\textit{Qwen2.5-Instruct-7B}}, to assess pattern reasoning ability without explicit temporal modeling;
(2) Specialized time series models, such as \textbf{\textit{FEDformer}}~\cite{zhou2022fedformer}, \textbf{\textit{PatchTST}}~\cite{nie2022time}, \textbf{\textit{TimesNet}}~\cite{wutimesnet}, and \textbf{\textit{iTransformer}}~\cite{liu2023itransformer}, which focus on modeling deterministic and probabilistic temporal patterns;
(3) TS-Text methods, including \textbf{\textit{GPT4TS}}~\cite{zhou2023one}, \textbf{\textit{Time-LLM}}~\cite{ICLR2024_680b2a81}, \textbf{\textit{ChatTS}}~\cite{xie2025chatts}, and \textbf{\textit{S\textsuperscript{2}IP}}~\cite{pan2024s}, which convert time series into text for pattern analysis and reasoning.

We conduct experiments on the TimerBed~\cite{liu2025picture}, which consists of six datasets spanning three categories. 

\textbf{\textit{Simple tasks (RCW, TEE)}} involve one-to-one pattern-to-label mappings, \textbf{\textit{Complex tasks (ECG, EMG)}} require integrating multiple  patterns, and \textbf{\textit{Probabilistic tasks (CTU, HAR)}} exhibit uncertain relationships between patterns and labels. 

\begin{table*}[t]
\centering
\caption{Accuracy (\%) on three pattern analysis task categories across six datasets.}
\resizebox{0.999\textwidth}{!}{
\renewcommand{\arraystretch}{0.9}
\begin{tabular}{c|c|cc|cc|cc|c}
\toprule
\midrule
\multirow{2}{*}{\textbf{Methods}} & \multirow{2}{*}{\textbf{Types}} 
& \multicolumn{2}{c|}{\textbf{Simple}} 
& \multicolumn{2}{c|}{\textbf{Complex}} 
& \multicolumn{2}{c|}{\textbf{Probabilistic}} 
& \multirow{2}{*}{\textbf{Weighted Avg.}} 
 \\
& &\textbf{RCW}& \textbf{TEE}& \textbf{ECG}& \textbf{EMG}& \textbf{CTU} &\textbf{HAR}\\
\midrule
\textbf{\textit{Random Guessing}} & -- &50.0&14.3&25.0&33.3&50.0&16.7&32.98 \\
\midrule
\multicolumn{9}{c}{\textbf{Zero-shot Test}} \\
\midrule
\rowcolor{lightyellow}\textbf{\textit{GPT-5.1}} & Proprietary API &60.9&24.8&27.1&39.5&52.7&36.2&40.34\\
\textbf{\textit{Qwen2.5-Instruct-7B}} &TS-Text&52.1&16.9&22.5&35.9&46.3&28.4&33.85 \\
\midrule
\multicolumn{9}{c}{\textbf{Full-shot Test}} \\
\midrule
\rowcolor{lightyellow}\textbf{\textit{Qwen2.5-Instruct-7B}} &TS-Text&67.8&40.5&28.8&44.9&59.1&52.0&45.65\\
\midrule
\textbf{\textit{FEDformer}} & TS &76.6&42.9&26.4&73.3&51.6&89.9&52.16\\
\rowcolor{lightyellow}\textbf{\textit{PatchTST}} & TS &82.1&57.1&24.8&60.0&64.0&79.6&52.09 \\
\textbf{\textit{TimesNet}} & TS &80.2&61.9&26.2&73.3&64.0&88.7&53.28 \\
\rowcolor{lightyellow}\textbf{\textit{iTransformer}} & TS &76.9&21.4&24.5&46.7&46.4&89.5&51.10 \\
\midrule
\textbf{\textit{GPT4TS}} & TS-Text &78.2&62.3&21.4&75.5&\textcolor{blue}{69.9}&87.6&50.01 \\
\rowcolor{lightyellow}\textbf{\textit{Time-LLM}} & TS-Text &62.5&47.3&25.6&63.8&60.2&88.4&46.61 \\
\textbf{\textit{ChatTS}} & TS-Text &71.3&50.8&28.4&69.4&52.6&77.5&49.81 \\
\rowcolor{lightyellow}\textbf{\textit{S\textsuperscript{2}IP}} & TS-Text &81.5&59.2&27.5&\textcolor{red}{82.7}&64.2&75.2&52.79 \\
\midrule
\textbf{\textit{TSAlign-3B}} &Ours&\textcolor{blue}{84.2}&\textcolor{blue}{71.4}&42.1&76.8&68.7&\textcolor{blue}{91.5}&\textcolor{blue}{63.28}\\
\rowcolor{lightyellow}\textbf{\textit{TSAlign-7B}} &Ours&\textcolor{red}{86.0}&\textcolor{red}{74.3}&\textcolor{red}{42.9}&\textcolor{blue}{81.7}&\textcolor{red}{77.4}&\textcolor{red}{93.2}&\textcolor{red}{64.60}\\
\midrule 
\bottomrule
\end{tabular}
}

\label{tab:main_results1}
\vspace{-0.5cm}
\end{table*}

\textbf{Main Results}.\quad 
As shown in Tab.~\ref{tab:main_results1}, TSAlign achieves the best overall performance across the six time series pattern analysis tasks. Specifically, for Simple and Complex tasks, TSAlign-7B attains the best results on all datasets, significantly outperforming both traditional time series models and existing TS-Text methods.
For the more challenging Probabilistic tasks, TSAlign also exhibits stronger robustness under uncertainty. In contrast, text-based methods generally underperform, suggesting that converting time series into discrete textual representations leads to information loss. 

\subsection{Ablation Study}
To evaluate the effectiveness of each component, we conduct a series of ablation studies: \textbf{\textit{representation granularity}}, \textbf{\textit{alignment strategy}}, and \textbf{\textit{fusion strategy}}. They assess the effects of representation design, alignment, and fusion. 

\noindent
\begin{minipage}[t]{0.57\linewidth}
\textbf{Representation Granularity}.\quad 
We analyze the effect of representation granularity by considering three modeling strategies: point-level, patch-level, and variable-level. As shown in Tab.~\ref{tab:granularity}, patch-level achieves the best performance across model scales and evaluation. QA denotes the QA reasoning task, and PA denotes the pattern analysis task.
\end{minipage}
\hfill
\begin{minipage}[t]{0.40\linewidth}
\centering
\vspace{-0.28cm}
\renewcommand{\arraystretch}{1.0}
\captionof{table}{Comparison of performance under different granularity levels.}
\vspace{-0.12cm}
\resizebox{\linewidth}{!}{
\begin{tabular}{c|cc|cc}
\toprule
\multirow{2}{*}{\textbf{Granularity}} 
& \multicolumn{2}{c|}{\textbf{TSAlign-3B}} 
& \multicolumn{2}{c}{\textbf{TSAlign-7B}} \\
& \textbf{QA (\%)} & \textbf{PA (\%)} & \textbf{QA (\%)} & \textbf{PA (\%)} \\
\midrule
\textbf{Patch-level (Ours)} & \textbf{79.4} & \textbf{63.3} & \textbf{85.1} & \textbf{64.6} \\
\textbf{Point-level} & 65.7 & 54.3 & 69.2 & 58.2 \\
\textbf{Variable-level} & 46.9 & 40.2 & 48.6 & 37.4 \\
\bottomrule
\end{tabular}
}
\label{tab:granularity}
\end{minipage}

\textbf{Alignment Strategy}.\quad  As shown in Tab.~\ref{tab:alignment_ablation}, alignment improves performance over the no-alignment baseline, highlighting the mismatch between time series and language representations. While prototype and attention-based methods leverage semantic structure, directly replacing the original features leads to information loss. Residual alignment mitigates this issue, and our gated formulation further improves performance by adaptively controlling semantic integration. Effective alignment requires semantic guidance while preserving and selectively integrating temporal information.

\textbf{Fusion Strategy}.\quad
As shown in Tab.~\ref{tab:fusion_ablation}, the fusion strategy plays a crucial role in integrating multi-dimensional information. Uniform fusion yields the lowest accuracy, as it treats all dimensions equally. Scalar-weight fusion provides moderate gains but remains limited in expressiveness. Softmax-based fusion further improves performance with adaptive weighting, yet its competitive normalization suppresses useful signals. In contrast, our gated fusion achieves the best results by allowing multiple informative dimensions to be preserved and combined.

\begin{table}[h]
\centering
\vspace{-0.3cm}
\begin{minipage}{0.49\linewidth}
\centering
\caption{Ablation on alignment strategies.}
\resizebox{\linewidth}{!}{
\begin{tabular}{c|c|cc|cc}
\toprule
\multirow{2}{*}{\textbf{Module}} 
& \multirow{2}{*}{\textbf{Strategy}}
& \multicolumn{2}{c|}{\textbf{TSAlign-3B}} 
& \multicolumn{2}{c}{\textbf{TSAlign-7B}} \\
& & \textbf{QA} & \textbf{PA} & \textbf{QA} & \textbf{PA} \\
\midrule
\multirow{6}{*}{\textbf{\makecell{Alignment\\Strategy}}} 
& Ours & \textbf{79.4} & \textbf{63.3} & \textbf{85.1} & \textbf{64.6} \\
& Residual Alignment & 75.3 & 56.4 & 81.2 & 59.7 \\
& Attention Alignment & 65.8 & 52.1 & 69.3 & 55.2 \\
& Prototype Matching & 63.8 & 54.4 & 65.4 & 56.7 \\
& Linear Alignment & 57.5 & 50.2 & 60.9 & 51.8 \\
& No Alignment & 52.3 & 46.1 & 55.7 & 48.0 \\
\bottomrule
\end{tabular}
}
\label{tab:alignment_ablation}
\end{minipage}
\hfill
\begin{minipage}{0.49\linewidth}
\centering
\renewcommand{\arraystretch}{1.29}
\caption{Ablation on multi-dimensional fusion.}
\resizebox{\linewidth}{!}{
\begin{tabular}{c|c|cc|cc}
\toprule
\multirow{2}{*}{\textbf{Module}} 
& \multirow{2}{*}{\textbf{Strategy}}
& \multicolumn{2}{c|}{\textbf{TSAlign-3B}} 
& \multicolumn{2}{c}{\textbf{TSAlign-7B}} \\
& & \textbf{QA} & \textbf{PA} & \textbf{QA} & \textbf{PA} \\
\midrule
\multirow{4}{*}{\textbf{\makecell{Fusion\\Strategy}}} 
& Ours & \textbf{79.4} & \textbf{63.3} & \textbf{85.1} & \textbf{64.6} \\
& Softmax Fusion & 66.3 & 54.8 & 73.5 & 56.0 \\
& Scalar-weight Fusion & 63.5 & 51.9 & 66.4 & 54.2 \\
& Uniform Fusion & 58.1 & 49.7 & 62.5 & 52.3 \\
\bottomrule
\end{tabular}
}
\label{tab:fusion_ablation}
\end{minipage}
\vspace{-0.2cm}
\end{table}
\subsection{Representation Analysis for Time Series–Language Space}
\begin{figure}[t]
    \centering
    
    \includegraphics[width=1\linewidth]{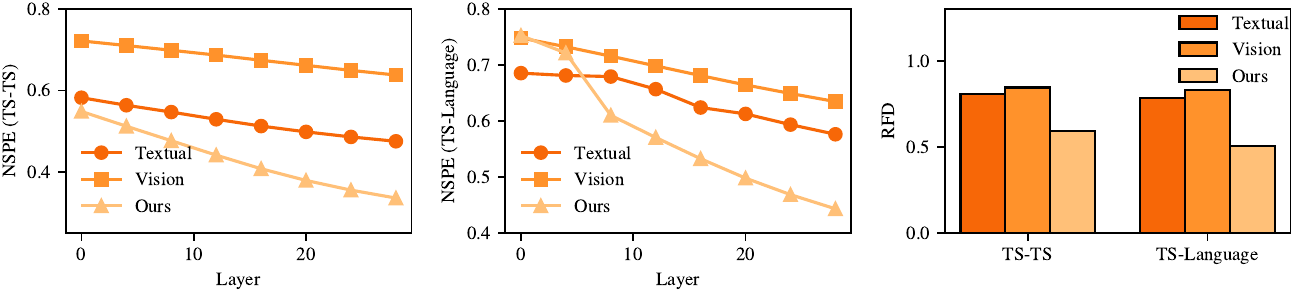}
    \caption{NSPE and RFD analysis of modality alignment. The left and middle plots show NSPE for TS–TS and TS–Language settings, respectively, while the right plot shows RFD.}
    \vspace{-0.3cm}
    \label{fig:nspe_rfd}
\end{figure}
\textbf{Normalized Subspace Projection Error (NSPE).}\quad
As shown in Fig.~\ref{fig:nspe_rfd}, NSPE ~\cite{mu2018all,ethayarajh2019contextual} measures the projection error of time series representations onto the principal subspace of language, where lower values indicate better alignment. TS–TS reflects intra-modal alignment within time series representations, while TS–Language measures cross-modal alignment. Under the TS–TS setting (left), as the LLM depth increases, our method achieves the lowest NSPE. Under the TS–Language setting (middle), our method reduces NSPE across all layers and continues to improve with depth, demonstrating its effectiveness in mitigating the modality gap. 

\textbf{Relative Fréchet Distance (RFD).}\quad
At the distribution level (right), we compute the Fréchet Distance~\cite{dowson1982frechet,heusel2017gans} based on the mean and covariance of representations and normalize it to obtain RFD, where lower values indicate better alignment. The results show that our method achieves the lowest RFD under both TS–TS and TS–Language settings, significantly outperforming all baselines.

\section{Conclusion \& Limitations}
In this work, we study LLM-based time series understanding beyond conventional state modeling. We introduce TSCognition, a multimodal benchmark for time series reasoning, which integrates time series with textual information and covers five cognitive reasoning tasks. Building on this, we propose TSAlign, a unified framework that aligns time series representations with the language reasoning space. Experiments on TSCognition and the TimerBed show that TSAlign achieves strong performance and efficiency, showing the value of language-conditioned time series reasoning.
This work still has several limitations. \textbf{\textit{First}}, TSCognition mainly adopts a multiple-choice format, which may not fully capture open-ended reasoning. \textbf{\textit{Second}}, although the dataset covers diverse domains, further expansion to more real-world scenarios and expert-level tasks would improve its coverage.
\clearpage
\bibliographystyle{abbrv}
\bibliography{Reference}

\appendix
\section{Ethics Statement}
Our study is limited to methodological and empirical investigation and does not involve human participants, animal subjects, or environmentally sensitive materials. We do not anticipate direct ethical concerns or conflicts of interest arising from this work. Nevertheless, since time-series models may be applied in high-stakes domains such as healthcare and critical industrial systems, their deployment should be conducted with caution, appropriate validation, and human oversight. All experiments and analyses are conducted with an emphasis on transparency, reproducibility, and responsible scientific practice.

\section{Limitation and Impact}
\subsection{Discussion of Limitations}
Although TSCognition and TSAlign achieve strong performance in time series language reasoning, this work still has several limitations. First, although TSCognition integrates real-world time series and textual information from 15 public sources and covers five cognitive reasoning tasks, namely Decoding, Grounding, Inferring, Extrapolating, and Acting, the current dataset still cannot exhaustively cover all complex time series scenarios in the real world. In high-stakes domains such as healthcare, industrial control, and financial risk management, temporal dynamics are often jointly shaped by latent mechanisms, external interventions, domain-specific rules, and expert knowledge, which are not yet fully captured by the current benchmark.

Second, this work mainly adopts multiple-choice QA as a unified evaluation format. This setting enables stable, reproducible, and quantifiable comparison, but it also limits the evaluation of open-ended explanation, generative reasoning, multi-turn interaction, and sequential decision-making. Future work may extend TSCognition to more complex task formats, such as free-form answers, step-by-step explanations, and interactive decision-making.

In addition, TSAlign still relies on pretrained LLMs for language understanding and reasoning, and its performance is therefore constrained by the capability boundary of the underlying LLM. Although semantic alignment, residual preservation, and multi-dimensional fusion help mitigate the modality gap between time series and language space, these mechanisms cannot fully eliminate potentially incorrect reasoning or unreliable explanations generated by LLMs. Meanwhile, TSAlign adopts patch-level representations to model local temporal structures and reduce computational cost. However, for irregularly sampled series, event-driven sequences, or highly noisy data, the current encoding strategy may still be suboptimal.

Finally, although we evaluate the generalization ability of TSAlign under both ID and OOD settings, real-world applications may involve more complex distribution shifts, missing observations, textual noise, and drifting task objectives. Future work can further expand the data types, task formats, and robustness evaluations to build more reliable time series language reasoning systems.

\subsection{Impact of Our Work}
The impact of this work lies in promoting a shift in time series analysis from conventional fixed-output modeling toward a reasoning-oriented paradigm centered on language, context, and decision objectives. By integrating time series signals, textual background, and natural language questions, TSCognition provides a systematic testbed for evaluating whether models can understand temporal dynamics, reason with semantic context, and support task-oriented judgment. Compared with traditional tasks that primarily focus on forecasting error or classification accuracy, this setting encourages the development of more interpretable time series intelligence systems that are closer to real-world analytical needs.

TSAlign has broad potential applications, including system monitoring, traffic management, energy scheduling, financial analysis, climate research, industrial operation and maintenance, and healthcare-assisted analysis. In these domains, users often require more than numerical predictions. They need models to understand the causes of anomalies, explain trend changes, incorporate contextual information, and support downstream decisions. By connecting multidimensional time series to LLM reasoning with lower token complexity and computational cost, TSAlign may reduce the barrier to applying LLM-based time series reasoning in long-sequence, multivariate, and real-time analysis scenarios.
At the same time, such methods should be used with caution. time series reasoning models may produce incorrect judgments when the input data contain noise, missing values, bias, or severe distribution shifts. LLMs may also generate explanations that appear plausible but are not reliable. Therefore, in high-stakes domains such as healthcare, finance, and industrial safety, TSAlign should not be regarded as an automated substitute for expert decision-making. Instead, it should be used as an auxiliary tool for analysis and decision support. Practical deployment should be combined with human review, domain knowledge, confidence estimation, and safety verification mechanisms to reduce the risks caused by erroneous reasoning.

\section{TSCognition Construction Pipeline}
\label{app:pipeline}
To construct high-quality time series reasoning samples, we adopt a unified QA construction pipeline consisting of four stages: raw data collection, data cleaning, QA construction, and quality control. 

\begin{figure}[t]
\begin{tcolorbox}[
    title=\textbf{QA Construction Pipeline of TSCognition},
    colback=gray!4,
    colframe=gray!65,
    boxrule=0.6pt,
    arc=2pt,
    left=4pt,
    right=4pt,
    top=4pt,
    bottom=4pt
]
\textbf{Step 1: Source Screening.}\quad

Select public sources containing both time series signals and semantically related textual information.

\textbf{Step 2: Data Cleaning.}\quad

Remove samples with missing values, timestamp errors, invalid variables, duplicates, noisy text, or mismatched time series--text pairs.

\textbf{Step 3: Unit Construction.}\quad

Organize each retained sample into a structured time series--text unit with temporal segments, variable descriptions, context, and metadata.

\textbf{Step 4: QA Sampling.}\quad

Construct questions following five tasks: Decoding, Grounding, Inferring, Extrapolating, and Acting.

\textbf{Step 5: Distractor Design.}\quad

Create plausible distractors from misleading local patterns, semantic mismatches, variable confusions, trend misreadings, or unreasonable extrapolations.

\textbf{Step 6: Human Verification.}\quad

Check answer uniqueness, question clarity, evidence support, and consistency among time series, text, answers, and distractors.

\textbf{Step 7: Final Assembly.}\quad

Keep only validated samples and standardize fields for benchmark construction and evaluation.
\end{tcolorbox}
\caption{QA construction pipeline of TSCognition}
\label{fig:pipeline}
\end{figure}
\textbf{Data Collection}.\quad During raw data collection, we prioritize data sources that contain both time series signals and relevant textual information, where the text may include event descriptions, variable definitions, scenario background, domain annotations, or task context. This design is not intended to simply concatenate time series and text in a superficial manner. Instead, it aims to ensure a clear semantic dependency between the two modalities, so that the generated questions can jointly evaluate the model’s understanding of temporal dynamics, textual context, and task objectives.

\textbf{Data Cleaning}.\quad During the data cleaning stage, we apply a unified preprocessing procedure to data from different sources, in order to reduce the impact of noise and format inconsistency on evaluation. Specifically, we remove samples with severe missing values, timestamp errors, abnormal temporal ordering, invalid variable fields, duplicate records, noisy textual descriptions, or mismatches between time series and text. For the retained data, we further organize the time series segments, variable information, textual background, and metadata into structured time series--text units, providing a unified input format for subsequent QA construction. This step enables data from different domains, sampling frequencies, and variable structures to be incorporated into a common evaluation framework.

\textbf{QA Construction}.\quad During the QA construction stage, we build questions around five cognitive reasoning tasks: Decoding, Grounding, Inferring, Extrapolating, and Acting. These tasks correspond to different reasoning requirements. Decoding focuses on identifying trends, periodicity, abrupt changes, and local patterns. Grounding evaluates the matching between time series patterns and textual context. Inferring emphasizes variable relationships, potential causes, and conditional reasoning. Extrapolating requires the model to infer future changes based on observed dynamics. Acting further requires the model to combine temporal evidence with task objectives to make decision-oriented judgments. For each question, we construct a unique correct answer and design challenging distractors. These distractors are mainly derived from time segments with similar local patterns but different conclusions, semantically plausible descriptions that are inconsistent with the data, common trend misinterpretations, confusion about variable relationships, or unreasonable extrapolations. This design prevents the questions from being overly superficial and requires the model to perform joint reasoning over both time series and textual information.

\textbf{Quality Control}.\quad During the quality control stage, each sample undergoes multiple rounds of human verification. Annotators first independently examine the consistency among the time series, textual background, question, correct answer, and distractors, checking whether the question is ambiguous, whether the answer is unique, whether the distractors are reasonable, and whether the reasoning process can be supported by the given information. For samples with disagreement, we introduce an additional reviewer for adjudication. Samples that cannot reach consensus or fail to satisfy the quality criteria are discarded. Through this multi-round verification mechanism, we reduce the influence of annotation bias, semantic ambiguity, and data noise on the benchmark, thereby improving the reliability and reproducibility of TSCognition across different domains and task types.

\section{Dataset Details}
\subsection{Case Study: Limitations of Existing TS Reasoning Datasets}
\label{app:Case Study}
To more concretely illustrate the limitations of existing  reasoning datasets, we conduct case studies on several representative question formats. \textbf{\textit{Cases 1--9 are collected from prior benchmarks}}, including Time-MQA~\cite{kong2025time} and Time-Omni-1~\cite{guan2025timeomni}, and are used here only for qualitative analysis. These examples highlight common limitations of existing datasets, such as simplified question formulations, limited use of textual context, weak multi-step reasoning requirements, and insufficient coverage of decision-oriented temporal tasks.

\begin{tcolorbox}[
    colback=gray!5,
    colframe=gray!60,
    boxrule=0.6pt,
    arc=2pt,
    left=4pt,
    right=4pt,
    top=4pt,
    bottom=4pt,
    title=\textbf{Case 1},
    fonttitle=\bfseries
]
\textbf{Question:} 

Which answer is the most comprehensive in relation to the question? Considering the data points 
[47.0, 51.0, 51.0, 20.0, 48.0, 44.0, 41.0, 28.0, 48.0, 38.0, 67.0, 66.0, 70.0,
54.0, 63.0, 56.0, 76.0, 58.0, 60.0, 58.0, 69.0, 56.0, 39.0, 50.0],
which statement best describes the volatility of this time series?

\vspace{0.3em}
\textbf{Candidate answers:} 

A) The volatility is low with consistent data points. 

B) There is moderate volatility with regular patterns. 

C) High volatility with sudden peaks and drops. 

D) The time series exhibits a perfectly linear trend.
\end{tcolorbox}

\begin{tcolorbox}[
    colback=gray!5,
    colframe=gray!60,
    boxrule=0.6pt,
    arc=2pt,
    left=4pt,
    right=4pt,
    top=4pt,
    bottom=4pt,
    title=\textbf{Case 2},
    fonttitle=\bfseries
]
\textbf{Question:} 

How clear or understandable is the reasoning provided in the answer? Considering the data points 
[60.0, 28.0, 46.0, 51.0, 37.0, 46.0, 41.0, 35.0, 32.0, 66.0, 10.0, 21.0, 67.0,
72.0, 83.0, 61.0, 11.0, 31.0, 62.0, 50.0, 56.0, 76.0, 53.0, 86.0],
do you see any seasonal patterns?

\vspace{0.3em}
\textbf{Candidate answers:} 

A) Yes, there is a seasonal pattern. 

B) No, there is no identifiable seasonal pattern.
\end{tcolorbox}

\begin{tcolorbox}[
    colback=gray!5,
    colframe=gray!60,
    boxrule=0.6pt,
    arc=2pt,
    left=4pt,
    right=4pt,
    top=4pt,
    bottom=4pt,
    title=\textbf{Case 3},
    fonttitle=\bfseries
]
\textbf{Question:} 

How clearly does the response explain its reasoning? (1 = Not Clear At All, 5 = Really Clear) 
What is the volatility level in the dataset 
[0.83, 0.8, 0.76, 0.75, 0.75, 0.76, 0.78, 0.8, 0.8, 0.8, 0.81, 0.8, 0.8, 0.8, 0.8, 0.79, 0.8, 0.82, 0.83, 0.82, 0.82, 0.82, 0.82, 0.81]?

\vspace{0.3em}
\textbf{Candidate answers:} 

A) High 

B) Moderate 

C) Low 

D) No volatility
\end{tcolorbox}

\begin{tcolorbox}[
    colback=gray!5,
    colframe=gray!60,
    boxrule=0.6pt,
    arc=2pt,
    left=4pt,
    right=4pt,
    top=4pt,
    bottom=4pt,
    title=\textbf{Case 4},
    fonttitle=\bfseries
]
\textbf{Question:} 

You manage a home energy storage system with battery capacity 18 kWh, current state-of-charge (SoC) 5 kWh, max charging power 5 kW, and max discharging power 10 kW. Historical 48-hour hourly load (kWh) from 2023-07-10 00:00 to 2023-07-11 23:00 is given. Tomorrow's 24-hour pricing is off-peak \$0.22/kWh for hours 0--14 and 20--23, and peak \$0.54/kWh for hours 15--19. Based on historical usage, predict tomorrow's load and select the optimal 24-hour battery strategy from A--D, where hour indices range from 0 to 23.

\vspace{0.3em}
\textbf{Candidate answers:} 

A) Charge: \{5, 11\}; Discharge: \{2, 18\}

B) Charge: \{1, 2\}; Discharge: \{15, 17\}

C) Charge: \{2, 10\}; Discharge: \{18, 19\}

D) Charge: \{11, 14\}; Discharge: \{16, 18\}
\end{tcolorbox}

\begin{tcolorbox}[
    colback=gray!5,
    colframe=gray!60,
    boxrule=0.6pt,
    arc=2pt,
    left=4pt,
    right=4pt,
    top=4pt,
    bottom=4pt,
    title=\textbf{Case 5},
    fonttitle=\bfseries
]
\textbf{Question:} 

You are given two time series related to river discharge measurements, expressed in m$^3$/s. Through causal discovery methods, we aim to identify potential causal relationships between different measuring stations from  data alone.

The time series of A463 is:
{\small
[4.12, 3.8, 3.44, 3.11, 2.97, 2.85, 2.79, 2.76, 2.63, 2.74, 2.69, 2.77, 2.75, 2.79, 2.97, 3, 3.09, 3.18, 3.24, 3.16, 3.16, 3.37, 3.37, 3.54, 3.56, 3.51, 3.43, 3.37, 3.37, 3.21, 3.12, 3.07, 3.16, 3.12, 3.16, 3.15, 2.95, 3, 2.95, 2.84, 2.75, 2.74, 2.74, 2.74, 2.74, 2.74, 2.6, 2.56, 2.56, 2.53, 2.54, 2.52, 2.33, 2.33, 2.33, 2.4, 2.61, 2.9, 3.22, 3.45, 3.58, 3.95]
} 

The time series of 4PRY is:
{\small
[7.55, 7.31, 7, 6.78, 6.5, 6.2, 6.08, 6, 6, 5.91, 5.82, 5.78, 5.84, 5.92, 5.88, 5.89, 5.82, 6, 6.25, 6, 5.94, 5.9, 5.9, 5.89, 5.89, 6, 6, 5.9, 5.74, 5.81, 5.67, 5.55, 5.35, 5.29, 5.14, 5.11, 5.18, 5.12, 5.11, 5.07, 4.89, 4.81, 4.82, 4.7, 4.69, 4.61, 4.67, 4.58, 4.45, 4.55, 4.42, 4.44, 4.32, 4.28, 4.37, 4.53, 4.5, 4.62, 4.84, 5, 5.43, 5.74]
}
Please identify the causal relationships between the two measurement stations. The data is collected every 12 hours from 2020-01-01 to 2020-01-31, with 62 points for each series.

\vspace{0.3em}
\textbf{Candidate answers:} 

A) A463 is the cause and 4PRY is the effect.

B) A463 and 4PRY are not causal.

C) 4PRY is the cause and A463 is the effect.
\end{tcolorbox}

\begin{tcolorbox}[
    colback=gray!5,
    colframe=gray!60,
    boxrule=0.6pt,
    arc=2pt,
    left=4pt,
    right=4pt,
    top=4pt,
    bottom=4pt,
    title=\textbf{Case 6},
    fonttitle=\bfseries
]
\textbf{Question:} 

You are given a time series. Please identify the scenario that most likely created it.
{\small
[39, 58, 51, 45, 40, 39, 57, 45, 45, 34, 40, 32.64, 62, 92, 37, 31, 29.52, 46, 69, 79, 43, 51, 50, 50, 56, 45, 57, 32, 35, 55, 38, 41, 38, 54, 33, 48, 54, 40.45, 30.31, 56, 49, 37.41, 45, 39, 49, 31.52, 47, 51, 33, 53, 38.49, 33, 30, 35.52, 41, 32.61, 52, 44, 29.68, 41, 32.44, 46.5, 49, 48, 38, 46, 38, 49, 45, 52, 47, 45, 39, 47, 44.47, 28, 36, 39, 27.67, 43, 37, 41, 52, 50, 37, 30, 30, 27.4, 33, 40, 35.51, 38, 35.43, 38, 37, 42, 28.49, 27, 47, 32.45, 30, 42, 36.39, 29.62, 27, 29.55, 32, 46, 44, 29.42, 26.72, 32, 44, 34.61, 37, 45, 45, 38.44, 42, 50, 34, 47.5, 51, 30, 31.56, 34, 42, 45, 39, 30, 42, 34, 50, 27.33, 38, 41.55, 32.43, 40, 40, 50, 39, 33.53, 26, 25.73, \ldots, 27, 30, 22.59, 38, 33, 26.58, 23.48, 32.62, 40, 20.33, 27.4, 33.44, 35.42, 33.34, 31, 38, 34, 21, 21, 38, 36]
}

\vspace{0.3em}
\textbf{Candidate answers:} 

A) ATM daily cash withdrawals over a year with an annual festival (365 daily samples).

B) Weekend music festival causing a surge in noise levels (96 hourly samples).

C) Hourly restaurant food orders during a 2-week local food festival (336 hourly samples).

D) NYC daily taxi pick-ups over a year with a subway strike (365 daily samples).
\end{tcolorbox}

\begin{tcolorbox}[
    colback=gray!5,
    colframe=gray!60,
    boxrule=0.6pt,
    arc=2pt,
    left=4pt,
    right=4pt,
    top=4pt,
    bottom=4pt,
    title=\textbf{Case 7},
    fonttitle=\bfseries
]
\textbf{Question:} 

You are given two time series related to river discharge measurements, expressed in m$^3$/s. Through causal discovery methods, we aim to identify potential causal relationships between different measuring stations from  data alone.

The time series of XV61 is:

{\small
[8.1, 7.15, 6.41, 6.42, 6.07, 7.76, 8.87, 7.5, 6.92, 6, 5.22, 5.43, 6.13, 6, 6.61, 5.85, 3, 2.83, 3.17, 4.11, 9.59, 16, 12.72, 8.45, 7, 6.3, 6, 5.74, 4.87, 4.59, 4.54, 4.39, 4.12, 3.91, 3.96, 3.83, 3.76, 3.8, 3.09, 3.1, 3.68, 3.83, 4.7, 4.39, 4.26, 4, 3.44, 3.33, 3.24, 3.22, 3.21, 3.16, 3, 3.04, 3.04, 3.04, 3, 3.04, 3.04, 3, 2.93, 2.79]
}

The time series of EJNN is:

{\small
[9.35, 9, 8.28, 7.49, 7.31, 7.89, 9.81, 9.7, 8.62, 7.82, 7, 6.35, 6.87, 6.8, 6.89, 7.12, 5.57, 4.17, 4.09, 4.51, 6.9, 13, 16.18, 13.56, 10.61, 8.44, 7.73, 7, 6.48, 5.38, 5.55, 5.74, 5.51, 5.08, 5, 4.95, 4.9, 4.74, 4.54, 3.9, 4.35, 5.07, 5.45, 5.43, 5.08, 5, 4.67, 4.05, 4.23, 4, 4, 3.9, 3.75, 3.53, 3.65, 3.56, 3.58, 3.54, 3.65, 3.63, 3.61, 3.57]
}

Please identify the causal relationships between the two measurement stations. The data is collected every 12 hours from 2020-03-01 to 2020-03-31, with 62 points for each series.

\vspace{0.3em}
\textbf{Candidate answers:} 

A) XV61 and EJNN are not causal.

B) EJNN is the cause and XV61 is the effect.

C) XV61 is the cause and EJNN is the effect.
\end{tcolorbox}

\begin{tcolorbox}[
    colback=gray!5,
    colframe=gray!60,
    boxrule=0.6pt,
    arc=2pt,
    left=4pt,
    right=4pt,
    top=4pt,
    bottom=4pt,
    title=\textbf{Case 8},
    fonttitle=\bfseries
]
\textbf{Question:} 

You are given two time series related to river discharge measurements, expressed in m$^3$/s. Through causal discovery methods, we aim to identify potential causal relationships between different measuring stations from  data alone.

The time series of 8S9U is:

{\small
[4.89, 7.75, 6.45, 6, 5.41, 6.81, 6.66, 5.52, 5, 4.86, 7.18, 6.64, 6, 6.13, 6.29, 5.74, 5.5, 5.83, 6.65, 6.07, 7.86, 7.2, 6.28, 5.61, 5.17, 5, 4.77, 4.53, 4.34, 4.17, 4.04, 3.83, 3.78, 3.65, 3.64, 3.5, 3.47, 3.47, 3.46, 3.42, 3.48, 3.34, 3.41, 3.31, 3.26, 3.22, 3.09, 3.05, 3.06, 3.04, 3.06, 3, 3, 2.94, 3, 2.92, 2.87, 2.85, 3.49, 3.42, 3.22, 3.07]
}

The time series of HUQ4 is:

{\small
[8, 10.32, 10.37, 9.4, 9, 9.87, 11, 9.32, 8.6, 8.4, 11.28, 11.87, 10.17, 10, 10.15, 9.76, 9.16, 9.28, 11, 10.67, 12.16, 12.57, 11, 10, 9, 8.34, 8.39, 7.83, 7.43, 7.2, 7, 6.79, 6.65, 6.36, 6.24, 6.19, 6.13, 6.08, 5.88, 5.82, 6, 6.13, 6.3, 6, 5.78, 5.61, 5.61, 5.72, 5.45, 5.42, 5.36, 5.49, 5.45, 5.44, 5.35, 5.23, 5.2, 5.3, 5.67, 6, 5.81, 5.54]
}

Please identify the causal relationships between the two measurement stations. The data is collected every 12 hours from 2020-03-01 to 2020-03-31, with 62 points for each series.

\vspace{0.3em}
\textbf{Candidate answers:} 

A) HUQ4 is the cause and 8S9U is the effect.

B) 8S9U and HUQ4 are not causal.

C) 8S9U is the cause and HUQ4 is the effect.
\end{tcolorbox}

\begin{tcolorbox}[
    colback=gray!5,
    colframe=gray!60,
    boxrule=0.6pt,
    arc=2pt,
    left=4pt,
    right=4pt,
    top=4pt,
    bottom=4pt,
    title=\textbf{Case 9},
    fonttitle=\bfseries
]
\textbf{Question:} 
You are given two time series related to river discharge measurements, expressed in m$^3$/s. Through causal discovery methods, we aim to identify potential causal relationships between different measuring stations from  data alone.

The time series of 994O is:

{\small
[41, 41, 41, 43, 53, 67, 72, 73, 74, 74, 73, 72, 71, 70, 69, 68, 67, 66, 65, 65, 63, 62, 61, 59, 58, 57, 57, 56, 56, 57, 56, 56, 55, 55, 53, 52, 52, 51, 50, 49, 48, 48, 47, 47, 46, 46, 45, 46, 47, 48, 48, 47, 47, 46, 46, 45.5, 45, 44, 44, 43, 43, 42]
}

The time series of XGMM is:

{\small
[42, 42, 42, 43, 52, 67, 72, 73, 74, 73, 72, 71, 70, 69, 67, 66, 65, 64, 63, 62, 60, 59, 58, 57, 55, 54, 53, 52, 52, 52, 52, 51, 50, 49, 48, 47, 46, 45, 44.44, 43.46, 42.47, 41.56, 41, 41, 40, 39.55, 39, 39, 40, 41, 41, 40, 40, 39.4, 39, 38, 37.5, 37, 36, 36, 35, 34.42]
}

Please identify the causal relationships between the two measurement stations. The data is collected every 12 hours from 2022-10-01 to 2022-10-31, with 62 points for each series.

\vspace{0.3em}
\textbf{Candidate answers:} 

A) 994O is the cause and XGMM is the effect.

B) 994O and XGMM are not causal.

C) XGMM is the cause and 994O is the effect.
\end{tcolorbox}

Overall, these cases suggest that existing time series reasoning datasets still have several limitations. Many questions mainly focus on shallow pattern recognition, such as detecting volatility, seasonality, or simple scenario matching, rather than requiring deeper reasoning over temporal dynamics. Some examples provide limited or ambiguous textual context, making it difficult to evaluate whether the model truly performs semantic grounding or merely relies on surface-level statistical cues. In addition, several tasks ask for causal, future-oriented, or decision-related judgments from time series data alone, but do not provide sufficient contextual evidence, domain constraints, or task objectives to support reliable reasoning. As a result, these benchmarks are useful for evaluating basic temporal pattern understanding, but remain limited in assessing language-conditioned  reasoning that requires joint modeling of signals, context, relations, and decisions.

\subsection{TimerBed}
\label{app:TimerBed}
\textbf{Overview.}\quad
TimerBed~\cite{liu2025picture} is a public benchmark for evaluating time series pattern analysis ability. In this work, we use TimerBed as the evaluation benchmark for the pattern analysis task. Unlike TSCognition, TimerBed does not require models to perform multi-step semantic reasoning over natural language questions, scenario context, and task objectives. Instead, it mainly evaluates whether models can identify discriminative temporal patterns from raw time series and map them to the corresponding categories. TimerBed categorizes time series pattern analysis tasks into three types: simple deterministic, complex deterministic, and probabilistic. Simple deterministic tasks mainly evaluate one-to-one correspondences between a single salient temporal pattern and its category. Complex deterministic tasks require models to integrate multiple temporal patterns for category prediction. Probabilistic tasks further introduce uncertain relationships between input patterns and labels, which are often influenced by unobserved user behaviors or external factors.

\textbf{Simple Deterministic.}\quad
Simple Deterministic consists of two tasks: \textbf{\textit{RCW}} and \textbf{\textit{TEE}}, where each category is mainly determined by a clear and localized temporal pattern. RCW, i.e., Right Whale Calls detection, is a North Atlantic right whale call detection task. It uses audio time series recorded in ocean environments to determine whether right whale calls are present, where a short rising ``whoop'' sound serves as the key discriminative pattern for the target class. 
TEE, i.e., Transient Electromagnetic Event classification, is derived from FORTE satellite observations and aims to identify transient electromagnetic events associated with lightning. Different event types correspond to relatively clear physical patterns in the power-density time series, such as a radiation spike followed by subsequent noise variations. 

\textbf{Complex Deterministic.}\quad
Complex Deterministic consists of two tasks: \textbf{\textit{ECG}} and \textbf{\textit{EMG}}, where each category is determined by the integration of multiple temporal patterns rather than a single local feature. ECG, i.e., Electrocardiogram Record Diagnosis, is a single-lead electrocardiogram diagnosis task. It aims to distinguish different cardiac conditions, including normal sinus rhythm, atrial fibrillation, other abnormal rhythms, and noisy or unclassifiable recordings, where the diagnosis usually depends on multiple signal characteristics such as rhythm irregularity, absence of P waves, QRS morphology, and baseline variations.
EMG, i.e., Electromyogram Signal Diagnosis, is a muscle-response diagnosis task based on electromyogram recordings under neural stimulation. It aims to classify signals into different physiological conditions, such as healthy, neuropathy, and myopathy, where the decision requires jointly considering multiple temporal features, including signal amplitude, duration, waveform complexity, and motor-unit response patterns.

\textbf{Probabilistic.}\quad
Probabilistic consists of two tasks: \textbf{\textit{HAR}} and \textbf{\textit{CTU}}, where the relationship between temporal patterns and labels is less deterministic and may be affected by unobserved factors. HAR, i.e., Human Activity Recognition, is a sensor-based activity classification task using smartphone recordings. It uses multichannel acceleration time series to identify daily activities such as walking, walking upstairs, walking downstairs, sitting, standing, and lying down, where user-specific behavior patterns may introduce uncertainty into the mapping between signals and labels.
CTU, i.e., Computer Type Usage classification, is a task that infers device type from 24-hour electricity usage time series. It aims to distinguish whether the usage pattern corresponds to a desktop or a laptop, where individual user habits, working schedules, and device usage behaviors can affect the observed temporal patterns and make the class relationship probabilistic.

\textbf{Statistics.}\quad
Tab.~\ref{tab:timerbed_stats} summarizes the statistics of the six sub-datasets. These datasets vary substantially in sample size, sequence length, and label space, ranging from small-scale datasets such as \textbf{\textit{TEE}} and \textbf{\textit{EMG}} to large-scale datasets such as \textbf{\textit{RCW}} and \textbf{\textit{ECG}}. Most datasets are univariate, while \textbf{\textit{HAR}} contains three variables, enabling evaluation under both single-variable and multivariate settings.

\begin{table}[h]
\centering
\caption{Statistics of the TimerBed datasets.}
\label{tab:timerbed_stats}
\resizebox{0.6\linewidth}{!}{
\begin{tabular}{c|cccc}
\toprule
\textbf{Dataset} & \textbf{ Variables} & \textbf{Time Series Length} & \textbf{ Classes} & \textbf{ Samples} \\
\midrule
\textbf{\textit{RCW}} & 1 & 4,000 & 2 & 30,000 \\
\textbf{\textit{TEE}} & 1 & 319  & 7 & 143 \\
\textbf{\textit{ECG}} & 1 & 1,500 & 4 & 43,673 \\
\textbf{\textit{EMG}} & 1 & 1,500 & 3 & 205 \\
\textbf{\textit{CTU}} & 1 & 720  & 2 & 500 \\
\textbf{\textit{HAR}} & 3 & 206  & 6 & 10,299 \\
\bottomrule
\end{tabular}
}
\end{table}

\subsection{TSCognition}
\label{app:TSCognition}
\textbf{Overview.}\quad
TSCognition is a multimodal time series language reasoning benchmark constructed in this work, aiming to evaluate whether models can jointly understand time series signals, textual context, and task objectives in real-world scenarios. Unlike pattern analysis benchmarks such as TimerBed, which mainly focus on temporal pattern classification, TSCognition emphasizes language-conditioned time series reasoning. That is, models are required not only to identify basic temporal patterns such as trends, periodicity, and abrupt changes, but also to integrate scenario descriptions, variable semantics, event backgrounds, and question objectives for semantic grounding, relational inference, future-oriented judgment, and decision analysis.
To this end, we collect real-world time series and their associated textual information from 15 public sources, resulting in approximately 41K QA samples. Each sample consists of a time series segment, textual background, a question, and candidate answers, requiring models to exploit numerical signals and language information to produce the correct answer.

\textbf{Hierarchical Cognitive Tasks.}\quad To characterize time series reasoning abilities from low-level pattern recognition to high-level decision-oriented reasoning, we design TSCognition around five hierarchical task categories: \textbf{\textit{Decoding, Grounding, Inferring, Extrapolating, and Acting}}. These categories are not isolated task types, but instead reflect a progressive process from “perceiving signals” to “understanding context,” and further to “inferring causes, anticipating changes, and supporting actions.”
Specifically, Decoding focuses on extracting basic temporal dynamics from raw time series, such as trends, periodicity, peaks, declines, abrupt changes, and local anomalies. Grounding further requires models to associate these temporal patterns with scenarios, events, or variable semantics described in the text. Inferring emphasizes relationships among variables, latent causes, and conditional reasoning. Extrapolating requires models to judge possible future changes based on observed dynamics and contextual information. Acting is oriented toward decision objectives, requiring models to translate time series evidence into reasonable operations, alerts, or intervention suggestions.

\textbf{\textit{Decoding is the most fundamental task level, mainly evaluating whether a model can accurately identify dynamic patterns directly presented in the time series itself.}} For example, in traffic flow data, the model may need to determine whether the vehicle flow on a road segment shows a sustained increase or a morning/evening rush-hour pattern. In server load data, it may need to identify whether CPU usage exhibits sudden spikes or a continuously increasing trend. In energy consumption data, it may need to judge whether the electricity usage curve shows a clear daily periodicity. In real-world applications, this ability corresponds to basic monitoring and anomaly detection, serving as the prerequisite for subsequent semantic understanding and decision analysis.

\textbf{\textit{Grounding introduces textual context, requiring models to determine whether the patterns observed in time series are consistent with the given scenario descriptions or event backgrounds.}} For example, when the text states that travel demand increases during a holiday period in a city, the model needs to judge whether the peaks in traffic flow align with this event. When the text describes that a factory enters a high-load production stage, the model needs to associate changes in energy consumption or equipment load with the production status. When the medical context indicates that a patient received an intervention during a specific period, the model needs to determine whether changes in physiological indicators can be reasonably linked to the intervention event. In real-world applications, this ability corresponds to contextualized monitoring and event interpretation: the model must not only observe signal changes, but also understand their meanings in specific scenarios.

\textbf{\textit{Inferring further requires models to perform relational reasoning based on both time series and textual information.}} Rather than merely determining whether a certain pattern exists, the model needs to analyze dependencies among variables, potential causes, and conditional changes. For example, in cloud system monitoring, if CPU usage and latency increase after a rise in request volume, the model needs to infer that the system bottleneck may come from computational resource pressure. In climate data, if temperature, humidity, and precipitation jointly exhibit abnormal changes, the model may need to identify the underlying weather-state transition. In financial scenarios, if price volatility and abnormal trading volume occur simultaneously, the model needs to judge whether market behavior has changed significantly. In real-world applications, this ability corresponds to root-cause analysis, relational modeling, and conditional diagnosis, serving as a key step from “observing phenomena” to “understanding mechanisms.”

\textbf{\textit{Extrapolating focuses on future-oriented judgment based on observed temporal dynamics.}} Unlike conventional forecasting, Extrapolating in TSCognition does not require models to output precise numerical predictions. Instead, models are expected to combine time series trends, contextual information, and candidate answers to identify the most likely future change. For example, in an energy load scenario, the model may need to determine whether a sustained increasing trend indicates a potential overload risk. In traffic scenarios, it may need to judge whether congestion in a certain area is likely to further deteriorate based on the current spreading pattern. In healthcare monitoring, the model may need to infer whether a patient’s condition is likely to continue worsening or become stable based on changes in vital signs. In real-world applications, this ability corresponds to early warning, trend assessment, and risk prediction. 

\textbf{\textit{Acting is the highest-level task category, requiring models to further translate temporal patterns, contextual understanding, relational inference, and future-oriented judgment into goal-oriented decisions.}} It evaluates not only whether a model knows “what has happened,” but also whether it can determine “what action should be taken.” For example, in cloud system operations, if the model identifies sustained increases in request volume, CPU usage, and latency, and further judges that service quality may degrade in the near future, a reasonable action may be to scale up computational resources or trigger load balancing. In traffic management, if congestion in a certain area continues to worsen, the model may need to select strategies such as traffic flow restriction, route diversion, or traffic signal scheduling. In industrial equipment monitoring, if vibration or temperature signals indicate potential failure risks, the model may need to recommend early maintenance or reduced operating load. In real-world applications, it corresponds to intelligent operations, decision support, and risk intervention, serving as a key step in moving time series language reasoning from understanding to practical application.

\begin{figure}[h]
    \centering
    \includegraphics[width=1\linewidth]{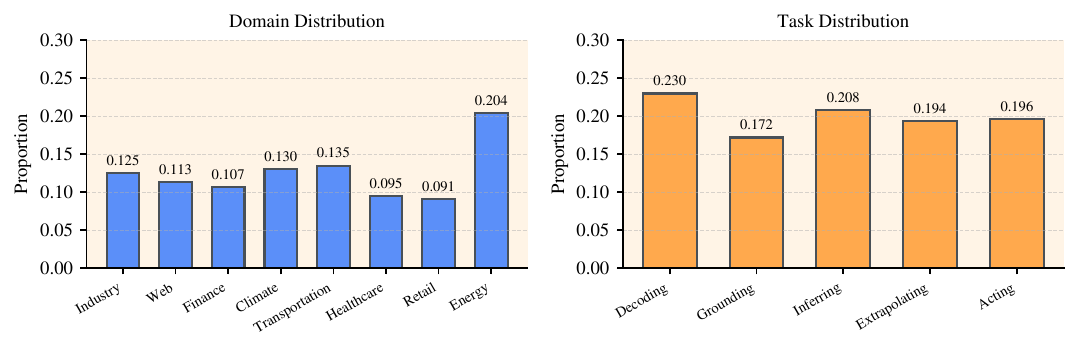}
    \caption{Distribution of TSCognition samples across eight domains and five hierarchical tasks.}
    \label{fig:dataset_distribution}
\end{figure}

\textbf{Statistics.}\quad
To provide a more intuitive overview of the data composition of TSCognition, we report the sample distribution from two perspectives: domain source and task type, as shown in Fig.~\ref{fig:dataset_distribution}. Overall, TSCognition covers diverse real-world domains and maintains a relatively balanced distribution across the five hierarchical reasoning tasks. This design ensures domain diversity while preventing the benchmark from being overly biased toward a specific reasoning type, thereby providing a more comprehensive basis for evaluating reasoning ability. In addition, Tab.~\ref{tab:task_level_statistics} reports task-level statistics of the time series, including the number of variables, sequence length, and answer choices.

\begin{table}[h]
\centering
\caption{Task-level statistics of time series in TSCognition.}
\label{tab:task_level_statistics}
\renewcommand{\arraystretch}{1.05}
\resizebox{0.95\linewidth}{!}{
\begin{tabular}{l|ccc|ccc|ccc}
\toprule
\multirow{2}{*}{\textbf{Tasks}} 
& \multicolumn{3}{c|}{\textbf{ Variable Numbers}} 
& \multicolumn{3}{c|}{\textbf{Time Series Length}} 
& \multicolumn{3}{c}{\textbf{ Choices}} \\
\cmidrule(lr){2-4} \cmidrule(lr){5-7} \cmidrule(lr){8-10}
& \textbf{Min} & \textbf{Max} & \textbf{Avg.}
& \textbf{Min} & \textbf{Max} & \textbf{Avg.}
& \textbf{Min} & \textbf{Max} & \textbf{Avg.} \\
\midrule
Decoding Task
& 4 & 5 & 4.7 
& 437 & 982  & 742 
& 4 & 5 & 4.7 \\

Grounding Task
& 3 & 6 & 4.4 
& 656 & 1018 & 871 
& 4 & 5 & 4.3 \\

Inferring Task
& 3 & 6 & 5.1 
& 882 & 1059 & 912 
& 4 & 5 & 4.6 \\

Extrapolating Task
& 3 & 6 & 5.2 
& 601 & 1052 & 846 
& 4 & 5 & 4.5 \\

Acting Task
& 4 & 6 & 5.5 
& 468 & 1047 & 795 
& 4 & 5 & 4.2 \\
\midrule
Overall 
& 3 & 6 & 4.9 
& 437 & 1059 & 785 
& 4 & 5 & 4.5 \\
\bottomrule
\end{tabular}
}
\end{table}
\section{Baselines}
\label{app:Baselines}
\textbf{GPT-5.1.}\quad
GPT-5.1 is the proprietary API baseline used in this work, representing a strong general-purpose LLM. We evaluate it under the zero-shot setting by organizing the time series information, textual background, question, and candidate answers into a unified natural language input. This baseline is used to examine the capability of a strong general LLM without task-specific training.

\textbf{Qwen2.5-Instruct~\cite{qwen2025qwen25technicalreport}.}\quad
Qwen2.5-Instruct is the large-scale open-source LLM baseline. We adopt the TS-Text input paradigm, where numerical time series values are textualized and then concatenated with the question, background description, and candidate answers as model input. This baseline is used to evaluate the reasoning ability of a large open-source LLM under textualized time series inputs, and to analyze whether increasing the LLM scale can alleviate the structural fragmentation and context burden caused by time series textualization.

\textbf{Qwen2.5-VL~\cite{qwen2025qwen25technicalreport}.}\quad
Qwen2.5-VL is the vision-language baseline. We convert time series into line plots and feed the resulting images, together with the textual background, question, and candidate answers, into the VLM. This baseline is used to evaluate the feasibility of reasoning over visualized time series with VLMs, and to examine whether visual representations can effectively preserve temporal structures and key dynamic patterns.

\textbf{ITFormer~\cite{wang2025itformer}.}\quad
ITFormer is a baseline designed for time series and language modeling, and is used to evaluate the performance of existing time series QA methods on TSCognition. This method typically connects time series information to LLMs through textualization or language interfaces, enabling the model to make predictions based on task descriptions and candidate answers. We include ITFormer to examine whether methods specifically designed for time series language tasks can adapt to more complex multi-level QA reasoning settings.

\textbf{Time-MQA~\cite{kong2025time}.}\quad
Time-MQA is a representative method for multimodal time series question answering. It combines time series with natural language questions to support QA tasks based on temporal dynamics and textual information. We use Time-MQA as a directly related time series QA baseline to compare existing QA construction paradigms with the more complex cognitive reasoning tasks in TSCognition.

\textbf{Time-LLM~\cite{ICLR2024_680b2a81}.}\quad
Time-LLM is a representative LLM-based time series method, which leverages LLM capabilities by converting or aligning time series into input forms that language models can process. We use it as one of the TS-Text baselines to evaluate the transferability of existing LLM4TS methods to QA reasoning scenarios. This comparison reflects whether LLM4TS methods originally designed for forecasting or representation learning remain effective under language-conditioned reasoning tasks.

\textbf{ChatTS~\cite{xie2025chatts}.}\quad
ChatTS is a baseline for time series understanding and conversational analysis. It typically represents time series in textual or LLM-processable forms, allowing the LLM to generate answers according to user questions. We include ChatTS to evaluate the performance of conversational time series analysis methods on standardized multiple-choice QA reasoning tasks.

\textbf{GPT4TS~\cite{zhou2023one}.}\quad
GPT4TS is an early representative method that introduces pretrained language models into time series modeling. It mainly leverages language-model architectures or pretrained knowledge to enhance time series representation learning, but its original design is more oriented toward conventional time series tasks rather than complex language-conditioned reasoning. We use GPT4TS as a baseline to analyze the applicability of such LLM-based time series methods to QA reasoning benchmarks such as TSCognition.

\textbf{S$^2$IP~\cite{pan2024s}.}\quad
S$^2$IP is adopted as a baseline for TimerBed, i.e., the pattern analysis task in our evaluation. It is designed to recognize characteristic temporal patterns from time series inputs rather than perform language-conditioned QA reasoning. Therefore, we use S$^2$IP to assess the capability of existing pattern analysis methods on the Simple Deterministic, Complex Deterministic, and Probabilistic categories.

\textbf{FEDformer~\cite{zhou2022fedformer}.}\quad
FEDformer is a frequency-enhanced time series modeling method that captures long-term dependencies and periodic structures through frequency-domain representations and decomposition mechanisms. In this work, we use FEDformer as a traditional TS baseline to evaluate the performance of non-LLM time series methods on the TimerBed pattern analysis tasks. This comparison helps determine whether the advantage of TSAlign comes only from LLM-based reasoning, or whether it also preserves effective modeling of raw temporal patterns.

\textbf{PatchTST~\cite{nie2022time}.}\quad
PatchTST is a patch-based Transformer model for time series modeling. It divides time series into local patches and performs representation learning at the patch level. Since TSAlign also adopts patch-level time series representations, PatchTST serves as an important reference for comparing pure time series patch modeling with patch representations aligned to the language space. This baseline is mainly used to examine the ability of patch-based TS models to recognize different types of temporal patterns in TimerBed without textual semantics or LLM reasoning.

\textbf{TimesNet~\cite{wutimesnet}.}\quad
TimesNet is a general-purpose time series analysis model that captures complex temporal variations through multi-period structure modeling. We use TimesNet as a strong time series baseline to evaluate the overall performance of traditional TS architectures on simple, complex, and probabilistic pattern analysis tasks.

\textbf{iTransformer~\cite{liu2023itransformer}.}\quad
iTransformer is an improved Transformer-based time series model that organizes tokens along the variable dimension to enhance multivariate time series modeling. In this work, iTransformer is used as a traditional TS baseline to evaluate the classification performance of non-textual time series models on TimerBed.

\clearpage
\section{Problem Formulation}
\label{app:Problem Formulation}
\subsection{Problem Formulation for QA Reasoning Tasks}
\label{app:qa_problem_formulation}

We first formalize the QA reasoning setting considered in TSCognition. Given a multivariate time series sample
$
X = \{x_1, x_2, \dots, x_T\},
$
where $x_t \in \mathbb{R}^d$, $T$ denotes the sequence length, and $d$ denotes the variable dimension. Unlike conventional time series tasks that predict fixed outputs such as future values, class labels, or anomaly indicators, QA reasoning requires the model to extract task-relevant temporal evidence and connect it with natural language semantics, scenario conditions, and decision objectives.

Let $q \in \mathcal{Q}$ denote a natural language query, which may contain task instructions, variable descriptions, contextual information, background events, or reasoning objectives. Let
$
\mathcal{A} = \{a_1, a_2, \dots, a_K\}
$
denote the set of candidate answers. The model is required to select the answer that is best supported by both the temporal evidence in $X$ and the semantic condition in $q$. This can be formulated as estimating the conditional distribution
$
p_\theta(a_k \mid X, q),
$
or equivalently learning a scoring function
$
s_k = S_\theta(X, q, a_k),
\label{eq:app_qa_score}
$
where $s_k$ measures the compatibility among the time series, the question, and the $k$-th candidate answer. The final prediction is given by
$
\hat{a}
=
\arg\max_{a_k \in \mathcal{A}} s_k.
\label{eq:app_qa_prediction}
$

To characterize the internal reasoning process, we introduce a candidate-conditioned latent state $z_k$, which represents the task-relevant temporal evidence induced by the triplet $(X, q, a_k)$. This state is not a generic time series embedding, but a candidate-specific representation that may contain local patterns, long-term trends, abrupt changes, cross-variable dependencies, and context-related temporal dynamics that support or refute the candidate answer. The conditional probability of each candidate can then be written as:
\begin{equation}
p_\theta(a_k \mid X, q)
=
\int p_\theta(a_k \mid z_k, q)\,
p_\theta(z_k \mid X, q, a_k)\, dz_k.
\label{eq:app_candidate_latent}
\end{equation}
Here, $p_\theta(z_k \mid X, q, a_k)$ describes how the model extracts candidate-specific temporal evidence from the input time series under the question condition, while $p_\theta(a_k \mid z_k, q)$ evaluates whether the candidate answer is semantically and temporally consistent with the inferred evidence.

This formulation highlights the conditional nature of QA reasoning. The same time series may require different evidence under different questions or candidate answers. For example, one question may focus on the global trend, another may focus on a local anomaly, while another may require reasoning over variable dependencies or future risks. Therefore, QA reasoning is not equivalent to directly classifying a time series into a fixed label space. Instead, it can be viewed as a task-conditioned evidence aggregation problem, where the model must dynamically select and organize temporal evidence according to the language query and candidate hypothesis.

During training, we adopt a multiple-choice negative log-likelihood objective. For each sample $(X, q, \mathcal{A}, a^\ast)$, where $a^\ast$ denotes the correct answer, the model first computes scores $\{s_1, s_2, \dots, s_K\}$ for all candidate answers. The conditional probability of each candidate is obtained through a softmax function:
\begin{equation}
P_\theta(a_k \mid X, q, \mathcal{A})
=
\frac{\exp(s_k)}
{\sum_{j=1}^{K}\exp(s_j)}.
\end{equation}
The training objective is then defined as
$
\mathcal{L}_{\mathrm{QA}}
=
-\log P_\theta(a^\ast \mid X, q, \mathcal{A}).
$
This objective requires the model to distinguish the correct answer from distractors by modeling the relative compatibility among temporal dynamics, textual context, and answer semantics. Since the candidate answer set varies across questions, the answer space is not fixed, making semantic alignment and conditional reasoning essential.

\subsection{Problem Formulation for Pattern Analysis Tasks}
\label{app:pa_problem_formulation}
We further formalize pattern analysis tasks, which are used to evaluate whether a model preserves fundamental temporal pattern modeling capability. Different from QA reasoning tasks, pattern analysis does not rely on an explicit natural language query. Given the same multivariate time series
$
X = \{x_1, x_2, \dots, x_T\},
x_t \in \mathbb{R}^d,
$
the model is expected to directly identify discriminative temporal structures from the raw time series and map them to a class label or pattern analysis target
$
y \in \mathcal{Y}.
$
This can be formulated as learning the conditional distribution
$
p_\theta(y \mid X),
$
or equivalently a mapping function
$
f_\theta: X \rightarrow y.
$

To describe the temporal pattern extraction process, we introduce a latent pattern representation $z$, which summarizes the task-relevant temporal structures contained in $X$. Such structures may include local peaks, periodic patterns, shape variations, temporal transitions, cross-variable coordination, or statistical dynamics. The prediction process can then be written as a marginalization over the latent pattern representation:
\begin{equation}
p_\theta(y \mid X)
=
\int p_\theta(y \mid z)\,
p_\theta(z \mid X)\, dz.
\end{equation}
Here, $p_\theta(z \mid X)$ extracts discriminative temporal patterns from the raw input, while $p_\theta(y \mid z)$ maps the inferred pattern representation to the target class.

The corresponding training objective is the standard supervised classification loss. For each sample $(X, y)$, the model maximizes the likelihood of the ground-truth label, or equivalently minimizes the negative log-likelihood:
$
\mathcal{L}_{\mathrm{PA}}
=
-\log p_\theta(y \mid X).
$
The core of pattern analysis is thus to learn the discriminative relationship between time series patterns and labels. It evaluates whether the model can capture useful temporal structures from raw signals, but does not require the model to interpret natural language questions or dynamically select evidence according to different semantic conditions.

\section{Prompt Template}
\label{app:Prompt}
To ensure comparability across different models and input paradigms, we use a unified prompt template to organize the time series, textual context, question, and candidate answers. For all LLM-based baselines, the prompt only provides the information required to complete the task, including the time-series content, variable descriptions, scenario background, question statement, and candidate options. It does not include any additional answer hints or manually designed reasoning chains. We organize each sample in a multiple-choice question answering format. The input first provides a task instruction, asking the model to select the most appropriate answer based on the time series and textual context. It then presents the variable descriptions, time-series segments, scenario background, and question, followed by the candidate answers. The model is required to output the corresponding option rather than generate open-ended long-form text. This design reduces evaluation instability caused by differences in generation style across models.

\textbf{TS-Text Baseline.}\quad
Time series values are converted into textual sequences using a unified format. For a univariate time series, the prompt presents the variable name together with its value list. For a multivariate time series, the observations of each variable are listed separately. This template preserves the temporal order and variable names, allowing the LLM to receive time-series information in textual form. The detailed prompt template is shown in Fig.~\ref{fig:prompt_template}.

\textbf{Vision-Text Baseline.}\quad
We render the time series as line plots and provide the same textual context, question, and candidate answers as in the QA reasoning setting. The model is required to answer by jointly considering the temporal patterns shown in the image and the accompanying textual information. To ensure fairness, all plots are generated using a unified visualization setup, including the same image size, axis format, variable annotation scheme, and color rules. The detailed prompt template is shown in Fig.~\ref{fig:prompt_template1}.

\textbf{TSAlign.}\quad
TSAlign does not fully textualize or visualize the time series. Instead, it feeds the temporal representation obtained after encoding, alignment, and fusion into the LLM together with the textual prompt. Therefore, at the prompt level, TSAlign uses the same question, context, and candidate-answer template as other QA methods; the only difference lies in how time-series information is incorporated. This ensures that the comparison focuses on the representation paradigm of time series itself, rather than differences in prompt wording. The detailed prompt template is shown in Fig.~\ref{fig:prompt_template2}.
\begin{figure}[h]
\begin{tcolorbox}[
    title=\textbf{Prompt Template for TS-Text Baseline},
    colback=mycyan,
    colframe=mycyan1,
    boxrule=0.6pt,
    arc=2pt,
    left=5pt,
    right=5pt,
    top=5pt,
    bottom=5pt
]
\textbf{Instruction:} You are an expert in time series analysis and temporal reasoning. Please answer the multiple-choice question based on the textualized time series, context, and candidate answers.

\textbf{Time Series:} \\
Variable 1: $[v^{1}_{1}, v^{1}_{2}, \dots, v^{1}_{T}]$ \\
Variable 2: $[v^{2}_{1}, v^{2}_{2}, \dots, v^{2}_{T}]$ \\
$\cdots$ \\
Variable $D$: $[v^{D}_{1}, v^{D}_{2}, \dots, v^{D}_{T}]$

\textbf{Variable Information:} [Variable names and descriptions]

\textbf{Context:} [Scenario background or task context]

\textbf{Question:} [Question]

\textbf{Candidate Answers:} \\
A. [Candidate answer A] \\
B. [Candidate answer B] \\
C. [Candidate answer C] \\
D. [Candidate answer D]

\textbf{Output:} Please output  the correct candidate answer.
\end{tcolorbox}
\caption{Prompt template for TS-Text baseline.}
\label{fig:prompt_template}
\end{figure}

\begin{figure}[h]
\begin{tcolorbox}[
    title=\textbf{Prompt Template for Vision-Text Baseline},
    colback=mycyan,
    colframe=mycyan1,
    boxrule=0.6pt,
    arc=2pt,
    left=5pt,
    right=5pt,
    top=5pt,
    bottom=5pt
]
\textbf{Instruction:} You are an expert in time series analysis and visual reasoning. Please answer the multiple-choice question based on the time series plot, context, and candidate answers.

\textbf{Time Series Image:} [Line plot of the time series]

\textbf{Variable Information:} [Variable names and descriptions]

\textbf{Context:} [Scenario background or task context]

\textbf{Question:} [Question]

\textbf{Candidate Answers:} \\
A. [Candidate answer A] \\
B. [Candidate answer B] \\
C. [Candidate answer C] \\
D. [Candidate answer D]

\textbf{Output:} Please output  the correct candidate answer.
\end{tcolorbox}
\caption{Prompt template for Vision-Text baseline.}
\label{fig:prompt_template1}
\end{figure}

\begin{figure}[h]
\begin{tcolorbox}[
    title=\textbf{Prompt Template for TSAlign},
    colback=mycyan,
    colframe=mycyan1,
    boxrule=0.6pt,
    arc=2pt,
    left=5pt,
    right=5pt,
    top=5pt,
    bottom=5pt
]
\textbf{Instruction:} You are an expert in time series analysis and temporal reasoning. A temporal token representing the input time series is provided to the model. Please answer the multiple-choice question based on the temporal evidence, context, and candidate answers.

\textbf{Temporal Token:} [Encoded time series representation inserted at the embedding level]

\textbf{Variable Information:} [Variable names and descriptions]

\textbf{Context:} [Scenario background or task context]

\textbf{Question:} [Question]

\textbf{Candidate Answers:} \\
A. [Candidate answer A] \\
B. [Candidate answer B] \\
C. [Candidate answer C] \\
D. [Candidate answer D]

\textbf{Output:} Please output  the correct candidate answer.
\end{tcolorbox}
\caption{Prompt template for TSAlign.}
\label{fig:prompt_template2}
\end{figure}
\section{Representation Analysis for Time Series--Language  Space}
\label{app:Representation}
To further examine whether TSAlign effectively mitigates the modality gap between time series and language representations, we analyze two distribution-level alignment metrics: Normalized Subspace Projection Error and Relative Fréchet Distance. These two metrics characterize the compatibility between time series representations and the language embedding space from complementary perspectives, namely geometric subspace alignment and distributional discrepancy. The motivation is that the input embedding space of a pretrained LLM is not an arbitrary high-dimensional vector space, but a structured semantic space shaped by large-scale language pretraining. Therefore, for time series representations to be effectively accepted and utilized by the LLM, matching the hidden dimensionality alone is insufficient; they should also approach the language representation space in both geometric orientation and distributional form.

\textbf{Normalized Subspace Projection Error (NSPE).}\quad
NSPE~\cite{mu2018all,ethayarajh2019contextual} measures the projection error of time series representations with respect to the principal language subspace. Specifically, we first take the word embedding matrix of the pretrained LLM and perform principal component analysis (PCA) on it to identify the dominant semantic directions in the language embedding space. These principal directions can be regarded as the dominant language subspace of the LLM input space. Intuitively, if most of the energy of a time series representation can be explained by this language subspace, then it is geometrically closer to the representations that the LLM is originally trained to process. In contrast, if a large portion of the representation lies outside this subspace, it indicates a more evident geometric mismatch with the language embedding space. Therefore, a lower NSPE suggests that the time series representation is better covered by the language principal subspace, indicating stronger language-space compatibility.

Formally, let $\mathbf{E} \in \mathbb{R}^{|\mathcal{V}| \times d}$ denote the word embedding matrix of the pretrained LLM, where $|\mathcal{V}|$ is the vocabulary size and $d$ is the embedding dimension. We perform PCA on $\mathbf{E}$ and take the top-$k$ principal components to form an orthonormal basis $\mathbf{U}_k \in \mathbb{R}^{d \times k}$. Given a time series representation $\mathbf{z} \in \mathbb{R}^{d}$, its projection onto the dominant language subspace is given by $\mathbf{U}_k \mathbf{U}_k^\top \mathbf{z}$, and the residual component not explained by this subspace is $\mathbf{z} - \mathbf{U}_k \mathbf{U}_k^\top \mathbf{z}$. We define NSPE as the ratio between the residual norm and the original representation norm:
\begin{equation}
\mathrm{NSPE}(\mathbf{z}) =
\frac{
\left\| \mathbf{z} - \mathbf{U}_k \mathbf{U}_k^\top \mathbf{z} \right\|_2
}{
\left\| \mathbf{z} \right\|_2
}.
\end{equation}

This normalized formulation removes the influence of representation scale, making comparisons across layers and methods more stable. A smaller $\mathrm{NSPE}(\mathbf{z})$ indicates that a larger proportion of $\mathbf{z}$ lies within the principal language subspace, whereas a larger value suggests that $\mathbf{z}$ contains more directions deviating from the language embedding space, which may make it harder for the LLM to effectively utilize the time series information. For a set of time series representations $\mathbf{Z}=\{\mathbf{z}_i\}_{i=1}^{n}$, we compute the average NSPE over samples:
$
\mathrm{NSPE}(\mathbf{Z}) =
\frac{1}{n}\sum_{i=1}^{n} \mathrm{NSPE}(\mathbf{z}_i),
$
which serves as an overall measure of geometric alignment at the corresponding layer.

In our experiments, we compute NSPE under two settings: TS--TS and TS--Language. The TS--TS setting mainly evaluates the internal consistency of time series representations, i.e., whether different time series samples gradually form a more stable and compact representation structure after being transformed through model layers. In contrast, TS--Language measures the cross-modal geometric alignment between time series representations and the principal language subspace, and thus serves as the core indicator for assessing whether the modality gap is mitigated.
As shown in Tab.~\ref{tab:nspe_results}, under the TS--TS setting, TSAlign achieves lower NSPE as the LLM depth increases, indicating that its time series representations progressively evolve into a more consistent internal structure. Under the TS--Language setting, TSAlign consistently reduces the projection error with respect to the principal language subspace across layers, with further improvements in deeper layers. This suggests that the proposed semantic alignment mechanism continuously alleviates the geometric mismatch between time series and language representations.
\begin{table}[h]
\centering
\caption{NSPE comparison under TS--TS and TS--Language settings across different LLM layers. Lower values indicate better alignment.}
\label{tab:nspe_results}
\renewcommand{\arraystretch}{1.0}
\resizebox{0.75\linewidth}{!}{
\begin{tabular}{c|ccc|ccc}
\toprule
\multirow{2}{*}{\textbf{Layer}} 
& \multicolumn{3}{c|}{\textbf{TS--TS}} 
& \multicolumn{3}{c}{\textbf{TS--Language}} \\
\cline{2-7}
& \textbf{Textual} & \textbf{Vision} & \textbf{Ours}
& \textbf{Textual} & \textbf{Vision} & \textbf{Ours} \\
\midrule
0  & 0.582 & 0.722 & 0.549 & 0.686 & 0.748 & 0.753 \\
4  & 0.564 & 0.710 & 0.512 & 0.681 & 0.733 & 0.721 \\
8  & 0.547 & 0.699 & 0.477 & 0.679 & 0.716 & 0.610 \\
12 & 0.529 & 0.687 & 0.442 & 0.657 & 0.698 & 0.571 \\
16 & 0.513 & 0.674 & 0.408 & 0.624 & 0.681 & 0.533 \\
20 & 0.499 & 0.662 & 0.380 & 0.613 & 0.665 & 0.498 \\
24 & 0.486 & 0.649 & 0.356 & 0.594 & 0.649 & 0.469 \\
28 & 0.476 & 0.638 & 0.336 & 0.577 & 0.635 & 0.444 \\
\bottomrule
\end{tabular}
}
\end{table}

\textbf{Relative Fréchet Distance (RFD).}\quad 
RFD~\cite{heusel2017gans,dowson1982frechet} measures the overall discrepancy between time series and language representations from a distributional perspective. Unlike NSPE, RFD further examines whether a set of time series representations is close to the language embedding distribution in terms of both mean and covariance structure. Therefore, NSPE focuses more on whether the representation directions are geometrically aligned, whereas RFD characterizes whether the overall representation distribution is distributionally compatible with the language space. We first use FD to measure the discrepancy between two representation distributions. Given two sets of representations with means $\boldsymbol{\mu}_1$ and $\boldsymbol{\mu}_2$, and covariance matrices $\boldsymbol{\Sigma}_1$ and $\boldsymbol{\Sigma}_2$, the FD is defined as:
\begin{equation}
\mathrm{FD}
=
\left\| \boldsymbol{\mu}_1 - \boldsymbol{\mu}_2 \right\|_2^2
+
\mathrm{Tr}\left(
\boldsymbol{\Sigma}_1 + \boldsymbol{\Sigma}_2
-
2\left(\boldsymbol{\Sigma}_1 \boldsymbol{\Sigma}_2\right)^{\frac{1}{2}}
\right).
\end{equation}
The first term measures the distance between the centers of the two distributions, while the second term captures the discrepancy between their covariance structures. Thus, a smaller FD indicates that the two representation sets are closer in both global location and distributional shape.
To enable relative comparison across different methods, we further define RFD. Let $\mathbf{Z}_m$ denote the set of time series representations produced by method $m$, and let $\mathbf{E}$ denote the set of language embeddings. The RFD of method $m$ with respect to a reference baseline is defined as:
\begin{equation}
\mathrm{RFD}(\mathbf{Z}_m, \mathbf{E})
=
\frac{
\mathrm{FD}(\mathbf{Z}_m, \mathbf{E})
}{
\mathrm{FD}(\mathbf{Z}_{\mathrm{ref}}, \mathbf{E})
}.
\end{equation}
Here, $\mathbf{Z}_{\mathrm{ref}}$ can be chosen as the unaligned time series representations or the representations produced by a basic baseline. Under this definition, $\mathrm{RFD}<1$ indicates that the current method is closer to the language embedding distribution than the reference baseline, whereas $\mathrm{RFD}>1$ indicates a larger distributional discrepancy. In our experiments, we mainly compare the relative distributional distances of different methods under the same setting; therefore, a lower RFD indicates better distribution-level alignment.

For each method, we extract time series representations from different LLM layers and compute their Fréchet Distance to the language embedding distribution. The TS--TS setting measures the relative distance among time series representation distributions, reflecting intra-modal consistency, while the TS--Language setting measures the distributional gap between time series and language embeddings. As shown in Tab.~\ref{tab:rfd_results}, TSAlign achieves the lowest RFD in both settings, with $0.594$ for TS--TS and $0.504$ for TS--Language, clearly outperforming Textual and Vision. This indicates that TSAlign forms more stable time series representations and brings their global distribution closer to the language embedding space, demonstrating stronger modality alignment.

\begin{table}[h]
\centering
\caption{RFD comparison under TS--TS and TS--Language settings. Lower values indicate better distributional alignment.}
\label{tab:rfd_results}
\renewcommand{\arraystretch}{1.15}
\resizebox{0.55\linewidth}{!}{
\begin{tabular}{l|ccc}
\toprule
\textbf{Setting} & \textbf{Textual} & \textbf{Vision} & \textbf{Ours} \\
\midrule
TS--TS       & 0.807 & 0.844 & 0.594 \\
TS--Language & 0.783 & 0.831 & 0.504 \\
\bottomrule
\end{tabular}
}
\end{table}

\section{Detailed Ablation Setting}
\label{app:ablation}
\subsection{Representation Granularity}
\textbf{Point-level.}\quad
In the point-level setting, we directly use raw timestamps as the basic modeling units. Given a multivariate time series $\mathbf{X}\in\mathbb{R}^{L\times D}$, the observation at each timestamp is treated as an independent temporal unit and projected into a representation space compatible with the LLM hidden size. This setting preserves the original temporal resolution to the greatest extent, but it also produces a long token sequence, which substantially increases the context length and computational cost, especially for long-horizon and multivariate time series. Moreover, an individual timestamp usually contains only local numerical information, without explicitly modeling short-term shapes, local trends, or local periodic structures.

\textbf{Variable-level.}\quad
In the variable-level setting, we first perform global aggregation over the complete time series of each variable, producing a single representation for each variable. In other words, all timestamps or patches of one variable are compressed into a global variable representation, and the representations of different variables are then fed into the subsequent fusion module. This strategy leads to shorter input length and lower computational cost, but it inevitably sacrifices fine-grained temporal structures, especially abrupt local changes, stage-wise variations, short-period patterns, and time-position-dependent information. Therefore, variable-level modeling can be viewed as a global summary-based strategy, which is used to examine whether excessive compression along the temporal dimension weakens the model's ability to understand complex temporal dynamics.

\textbf{Patch-level (Ours).}\quad
In the patch-level setting, we adopt the strategy used in our main model, where the time series of each variable is divided into a set of consecutive or overlapping local patches. Each patch covers a local temporal window and is mapped by the time-series encoder into a patch-level representation. This design provides a balance between raw point-level modeling and global variable-level summarization. On the one hand, it preserves local dynamic patterns, such as short-term trends, abrupt changes, periodic fragments, and local fluctuations. On the other hand, it significantly reduces the sequence length compared with point-level modeling, allowing the model to receive temporal information in a more compact form. In the subsequent modules, patch-level representations are further processed by language-space alignment and multivariate fusion, and are finally used as temporal representations for LLM reasoning.

\subsection{Alignment Strategy}
\textbf{No Alignment.}\quad
This setting removes the explicit alignment module between time-series representations and the language space. Specifically, the representations produced by the time-series encoder are only passed through a necessary linear projection to match the LLM hidden size, and are then directly used as temporal tokens for downstream reasoning. This baseline examines whether dimensional compatibility alone is sufficient for the LLM to effectively utilize time-series information without semantic alignment constraints.

\textbf{Linear Alignment.}\quad
This setting uses a learnable linear mapping to project time-series representations into the LLM embedding space. Compared with No Alignment, Linear Alignment introduces an additional parametric transformation, enabling the model to learn a direct mapping from the time-series feature space to the language representation space. However, it does not explicitly exploit the semantic structure encoded in pretrained word embeddings, and therefore can only learn cross-modal transformation implicitly from supervised signals.

\textbf{Prototype Matching.}\quad
This setting uses semantic prototypes extracted from the LLM word embedding space as alignment targets. Specifically, we construct a set of representative prototypes from pretrained language embeddings and map time-series representations into the semantic space spanned by these prototypes. This baseline evaluates whether explicitly introducing the structure of the language space helps mitigate the modality gap. However, its representation capacity is constrained by discrete prototypes, which may make it difficult to preserve complete and continuous temporal dynamics.

\textbf{Attention Alignment.}\quad
This setting takes time-series representations as queries, and uses language embeddings or their compressed semantic bases as keys and values to generate language-induced representations through an attention mechanism. Compared with Prototype Matching, Attention Alignment can adaptively select relevant semantic directions for each time-series representation, and therefore provides more flexible cross-modal alignment. This baseline examines whether directly generating language-space representations through semantic attention can effectively replace the original time-series features.

\textbf{Residual Alignment.}\quad
This setting adds a residual connection on top of Attention Alignment. Specifically, the model no longer directly replaces the original time-series representation with the language-induced representation, but instead treats it as a semantic correction term added to the original features. This baseline evaluates the importance of preserving raw time-series information during cross-modal alignment, preventing potential loss of temporal dynamics caused by direct semantic replacement.

\textbf{Ours Alignment.}\quad
This setting corresponds to the alignment strategy adopted in TSAlign. Building upon Residual Alignment, it further introduces a learnable gate to adaptively control the strength of injecting the semantic correction term into the original time-series representation. Specifically, when language semantics are beneficial for representing the current temporal segment or variable, the gate can enhance semantic information; when the original temporal dynamics are more important, the gate can suppress excessive semantic intervention. This design enables a dynamic balance between semantic guidance and temporal information preservation.

\subsection{Fusion Strategy}
\textbf{Uniform Fusion.}\quad
This setting directly averages the time-series representations of different variables. Specifically, each variable is assigned the same weight, without introducing additional learnable parameters or sample-adaptive mechanisms. This baseline examines whether simple averaging is sufficient to integrate multivariate time-series information when variable importance is not explicitly distinguished.

\textbf{Scalar-weight Fusion.}\quad
This setting introduces a learnable scalar weight for each variable. Compared with Uniform Fusion, Scalar-weight Fusion allows the model to learn the global importance of different variables during training, thereby assigning larger weights to more informative variables. However, these weights are typically static and cannot adapt to different samples or temporal dynamics, which limits their expressiveness.

\textbf{Softmax Fusion.}\quad
This setting assigns adaptive weights to different variables through softmax normalization. Specifically, the model computes a fusion weight from the representation of each variable and applies softmax to ensure that all variable weights sum to one. This baseline can dynamically adjust variable contributions according to the input sample. However, the competitive normalization of softmax may cause mutual suppression among variables; when multiple variables contain useful information simultaneously, some informative variables may be underestimated.

\textbf{Our Fusion.}\quad
This setting is the multivariate fusion strategy adopted in TSAlign. It independently computes a gate for each variable and controls its contribution to the final representation in a non-competitive manner. Unlike Softmax Fusion, Gated Fusion does not force all variables to compete for the same probability mass, and can therefore preserve information from multiple important variables at the same time. This design is more suitable for multivariate time-series scenarios, where multiple dimensions often provide complementary evidence rather than a single dominant variable determining the final prediction.
\section{Reproducibility}
\label{app:Reproducibility}
\subsection{Training Configuration}
To ensure reproducibility, we use unified implementation, training, and evaluation settings whenever possible. Unless otherwise specified, all experiments are conducted on two NVIDIA A800 GPUs. We use PyTorch as the main deep learning framework and implement the LLM backbone, tokenizer, and text input processing pipeline based on HuggingFace Transformers. For multi-GPU training and memory optimization, we use DeepSpeed and Accelerate to manage distributed training, ensuring that models with different scales and input paradigms can be trained and evaluated under the same hardware environment. 

All experiments are run with a fixed random seed of 2026 to control the effects of model initialization, data loading order, sample shuffling, dropout, and other stochastic operations. For each group of comparative experiments, we use the same training, validation, and test splits, and follow the same ID/OOD evaluation protocol, so that the performance differences among methods mainly arise from model design or input paradigm rather than random data partitioning or evaluation procedure differences.
All models are trained under a supervised learning paradigm. We adopt AdamW as the optimizer and keep the learning rate, weight decay, batch size, number of training epochs, and learning-rate scheduling strategy consistent within the same experimental group. For LLM-based methods, we use the official tokenizer and pretrained weights of the corresponding backbone, and maintain the same maximum input length, candidate-answer format, and prompt organization. At the implementation level, we use mixed-precision training to reduce memory consumption and improve training efficiency. More details are provided in Tab.~\ref{tab:implementation_details}.
\begin{table}[t]
\centering
\caption{Common experimental hyperparameters and training configuration.}
\label{tab:implementation_details}
\resizebox{0.80\linewidth}{!}{
\begin{tabular}{l|c}
\toprule
GPUs & 2 $\times$ NVIDIA A800 GPUs \\
Framework & PyTorch + HuggingFace Transformers \\
Distributed training & DeepSpeed / Accelerate \\
Precision & bfloat16 \\
Optimizer & AdamW \\
Learning rate & $2 \times 10^{-5}$ \\
Weight decay & $1 \times 10^{-2}$ \\
Batch size per GPU & 4 \\
Gradient accumulation & 4 \\
Effective batch size & 32 \\
Gradient clipping & 1.0 \\
Learning rate scheduler & Cosine decay \\
Warmup ratio & 0.0 \\
Random seed & 2026 \\
\bottomrule
\end{tabular}
}
\end{table}

\subsection{Implementation Details}
For QA reasoning tasks, we adopt a multiple-choice question answering format for both training and evaluation. Each sample consists of a time series, textual context, a question, and a set of candidate answers. The training objective is the standard cross-entropy loss. For the TS-Text baseline, we convert numerical time-series values into text using a unified template, and concatenate them with the question and candidate answers before feeding them into the LLM. For the Vision-Text baseline, we render the time series as line plots and feed the visual input together with the textual input into the VLM. For TSAlign, we first perform patch-based encoding over the time series, followed by language-space alignment and multivariate fusion. The resulting temporal representation is then combined with text tokens and fed into the LLM for joint reasoning.

For pattern analysis tasks, we follow the standard classification setting of TimerBed, where each time-series sample is mapped to its corresponding class label. For conventional time-series models, such as FEDformer, PatchTST, TimesNet, and iTransformer, we train and test them using their original raw time-series input format. For LLM-based or TS-Text methods, we adopt the same textualized input strategy as in QA reasoning. For TSAlign, we keep the time-series encoding, alignment, and fusion modules identical to those used in the main experiments, while replacing the output target with the corresponding pattern label. All methods are compared under the same training, validation, and test splits to ensure a fair evaluation.

In the implementation of TSAlign, each multivariate time series is first decomposed into patch-level representations. Each variable is independently processed by the time-series encoder to obtain local temporal representations, which are then passed through the language-space alignment module to introduce semantic guidance. A gated residual mechanism is used to preserve the original temporal dynamics while injecting language-induced information. For multivariate inputs, we use gated fusion to adaptively integrate different variables into a unified time-series representation. This representation is projected into the LLM hidden space and used as a temporal token together with textual inputs for reasoning. All ablation experiments are conducted within the same codebase, where only the module under study is replaced, such as representation granularity, alignment strategy, or fusion strategy, while all other training and evaluation settings remain unchanged.

\clearpage
\section{Examples of Cognitive Reasoning Tasks}
\label{app:cognitive_examples}

To illustrate the hierarchical design of TSCognition, we provide representative examples for the five cognitive reasoning tasks in Fig.~\ref{fig:prompt_template_decoding}--\ref{fig:prompt_template_acting}. These examples are constructed under the same regional power dispatch scenario, but require different levels of temporal understanding. The Decoding example focuses on identifying the basic pattern of renewable output, while Grounding requires matching temporal evidence with the operational context. Inferring further asks the model to explain possible causes behind system pressure, Extrapolating evaluates future-oriented judgment under evolving demand and renewable conditions, and Acting requires selecting an appropriate dispatch decision based on the observed temporal evidence. 

Together, these examples show how TSCognition moves beyond  pattern recognition and evaluates a progressive reasoning process from temporal perception to contextual understanding and decision-oriented action.

\begin{figure}[h]
\begin{tcolorbox}[
    title=\textbf{Example of the Decoding task},
    colback=mycyan,
    colframe=mycyan1,
    boxrule=0.6pt,
    arc=2pt,
    left=5pt,
    right=5pt,
    top=5pt,
    bottom=5pt
]
\textbf{Instruction:} You are an expert in time series analysis and temporal reasoning. A temporal token representing the input time series is provided to the model. Please answer the multiple-choice question based on the temporal evidence, context, and candidate answers. Only one answer is correct.

\textbf{Variable Information:} 
$D$: Load Demand; 
$G$: Thermal Generation; 
$R$: Renewable Output; 
$B$: Battery Power; 
$P$: Electricity Price; 
$F$: Frequency Deviation.

\textbf{Context:} 
The data describe the operation of a regional power system. 
The operator monitors load demand, thermal generation, renewable output, battery behavior, electricity price, and frequency deviation to evaluate supply-demand balance and dispatch decisions. 
Battery power reflects storage operation, while electricity price and frequency deviation may indicate system stress. 
The actual temporal state should be inferred from the provided time series representation.

\textbf{Question:} 
Which statement best characterizes the joint temporal behavior of renewable output and battery power?

\textbf{Candidate Answers:} \\
A. Renewable output stays low throughout the period, while the battery remains inactive with no clear charging or discharging phase. \\
B. Renewable output first increases to a mid-period peak and then declines, while the battery tends to charge during the renewable-rich phase and discharge later. \\
C. Renewable output and battery power increase monotonically together, indicating that storage output always follows renewable generation. \\
D. Renewable output reaches its highest level only during the final high-demand period, while the battery mainly charges at the same time. \\
E. Renewable output fluctuates randomly without a recognizable phase, while battery behavior shows no relation to the system state.

\textbf{Output:} Please output the correct candidate answer.
\end{tcolorbox}
\caption{Example of the Decoding task.}
\label{fig:prompt_template_decoding}
\end{figure}

\begin{figure}[h]
\begin{tcolorbox}[
    title=\textbf{Example of the Grounding task},
    colback=mycyan,
    colframe=mycyan1,
    boxrule=0.6pt,
    arc=2pt,
    left=5pt,
    right=5pt,
    top=5pt,
    bottom=5pt
]
\textbf{Instruction:} You are an expert in time series analysis and temporal reasoning. A temporal token representing the input time series is provided to the model. Please answer the multiple-choice question based on the temporal evidence, context, and candidate answers. Only one answer is correct.

\textbf{Variable Information:} 
$D$: Load Demand; 
$G$: Thermal Generation; 
$R$: Renewable Output; 
$B$: Battery Power; 
$P$: Electricity Price; 
$F$: Frequency Deviation.

\textbf{Context:} 
The data describe the operation of a regional power system. 
The operator monitors load demand, thermal generation, renewable output, battery behavior, electricity price, and frequency deviation to evaluate supply-demand balance and dispatch decisions. 
Battery power reflects storage operation, while electricity price and frequency deviation may indicate system stress. 
The actual temporal state should be inferred from the provided time series representation.

\textbf{Question:} 
Which interpretation best grounds the observed temporal patterns in the operational context of power dispatch?

\textbf{Candidate Answers:} \\
A. A renewable-rich period is reflected by higher renewable output together with battery charging, suggesting that surplus renewable energy is being absorbed. \\
B. A renewable-rich period is reflected by high electricity price and large frequency deviation, because these always indicate abundant renewable supply. \\
C. A peak-load period is reflected by low demand and strong battery charging, suggesting that the system is under severe supply shortage. \\
D. A stable operating period is reflected by simultaneous increases in load demand, electricity price, and frequency deviation, regardless of renewable output. \\
E. The storage behavior cannot be grounded in the power-system context because battery power is unrelated to renewable output or demand.

\textbf{Output:} Please output the correct candidate answer.
\end{tcolorbox}
\caption{Example of the Grounding task.}
\label{fig:prompt_template_grounding}
\end{figure}

\begin{figure}[h]
\begin{tcolorbox}[
    title=\textbf{Example of the Inferring task},
    colback=mycyan,
    colframe=mycyan1,
    boxrule=0.6pt,
    arc=2pt,
    left=5pt,
    right=5pt,
    top=5pt,
    bottom=5pt
]
\textbf{Instruction:} You are an expert in time series analysis and temporal reasoning. A temporal token representing the input time series is provided to the model. Please answer the multiple-choice question based on the temporal evidence, context, and candidate answers. Only one answer is correct.

\textbf{Variable Information:} 
$D$: Load Demand; 
$G$: Thermal Generation; 
$R$: Renewable Output; 
$B$: Battery Power; 
$P$: Electricity Price; 
$F$: Frequency Deviation.

\textbf{Context:} 
The data describe the operation of a regional power system. 
The operator monitors load demand, thermal generation, renewable output, battery behavior, electricity price, and frequency deviation to evaluate supply-demand balance and dispatch decisions. 
Battery power reflects storage operation, while electricity price and frequency deviation may indicate system stress. 
The actual temporal state should be inferred from the provided time series representation.

\textbf{Question:} 
What is the most plausible explanation for the system pressure indicated by electricity price and frequency deviation during the high-load phase?

\textbf{Candidate Answers:} \\
A. The pressure is mainly caused by decreasing demand, since lower load usually forces the system to increase electricity price and frequency deviation. \\
B. The pressure is likely caused by the coincidence of high load demand, reduced renewable support, and increased dependence on dispatchable resources. \\
C. The pressure is mainly caused by excessive renewable generation, because abundant renewable output always increases electricity price and frequency deviation. \\
D. The pressure is unrelated to supply-demand balance and is only caused by random fluctuations in the battery power curve. \\
E. The pressure disappears because battery discharge and thermal generation fully eliminate all frequency deviation.

\textbf{Output:} Please output the correct candidate answer.
\end{tcolorbox}
\caption{Example of the Inferring task.}
\label{fig:prompt_template_inferring}
\end{figure}

\begin{figure}[h]
\begin{tcolorbox}[
    title=\textbf{Example of the Extrapolating task},
    colback=mycyan,
    colframe=mycyan1,
    boxrule=0.6pt,
    arc=2pt,
    left=5pt,
    right=5pt,
    top=5pt,
    bottom=5pt
]
\textbf{Instruction:} You are an expert in time series analysis and temporal reasoning. A temporal token representing the input time series is provided to the model. Please answer the multiple-choice question based on the temporal evidence, context, and candidate answers. Only one answer is correct.

\textbf{Variable Information:} 
$D$: Load Demand; 
$G$: Thermal Generation; 
$R$: Renewable Output; 
$B$: Battery Power; 
$P$: Electricity Price; 
$F$: Frequency Deviation.

\textbf{Context:} 
The data describe the operation of a regional power system. 
The operator monitors load demand, thermal generation, renewable output, battery behavior, electricity price, and frequency deviation to evaluate supply-demand balance and dispatch decisions. 
Battery power reflects storage operation, while electricity price and frequency deviation may indicate system stress. 
The actual temporal state should be inferred from the provided time series representation.

\textbf{Question:} 
Assume that the current high-load phase continues for the next interval and renewable output does not recover. Which future evolution is most consistent with the observed temporal evidence?

\textbf{Candidate Answers:} \\
A. Electricity price and frequency deviation may further increase unless additional dispatchable generation or storage discharge is available. \\
B. Renewable output will necessarily rebound immediately, so electricity price and frequency deviation should both drop to their minimum values. \\
C. Battery power is more likely to switch to strong charging, because a high-load phase implies surplus electricity in the system. \\
D. Thermal generation will become unnecessary, since declining renewable output reduces the need for dispatchable resources. \\
E. The system will remain risk-free even if demand keeps increasing, because frequency deviation is independent of supply-demand imbalance.

\textbf{Output:} Please output the correct candidate answer.
\end{tcolorbox}
\caption{Example of the Extrapolating task.}
\label{fig:prompt_template_extrapolating}
\end{figure}

\begin{figure}[h]
\begin{tcolorbox}[
    title=\textbf{Example of the Acting task},
    colback=mycyan,
    colframe=mycyan1,
    boxrule=0.6pt,
    arc=2pt,
    left=5pt,
    right=5pt,
    top=5pt,
    bottom=5pt
]
\textbf{Instruction:} You are an expert in time series analysis and temporal reasoning. A temporal token representing the input time series is provided to the model. Please answer the multiple-choice question based on the temporal evidence, context, and candidate answers. Only one answer is correct.

\textbf{Variable Information:} 
$D$: Load Demand; 
$G$: Thermal Generation; 
$R$: Renewable Output; 
$B$: Battery Power; 
$P$: Electricity Price; 
$F$: Frequency Deviation.

\textbf{Context:} 
The data describe the operation of a regional power system. 
The operator monitors load demand, thermal generation, renewable output, battery behavior, electricity price, and frequency deviation to evaluate supply-demand balance and dispatch decisions. 
Battery power reflects storage operation, while electricity price and frequency deviation may indicate system stress. 
The actual temporal state should be inferred from the provided time series representation.

\textbf{Question:} 
Which operational decision best matches the observed system state and dispatch objective?

\textbf{Candidate Answers:} \\
A. Increase battery charging and reduce dispatchable generation, because the system appears to have large surplus supply during the high-load phase. \\
B. Maintain the current dispatch without intervention, because electricity price and frequency deviation provide no useful signal about operating pressure. \\
C. Discharge the battery and increase dispatchable generation to relieve the high-load pressure while monitoring frequency stability. \\
D. Shut down renewable generation completely, because removing renewable output is the most direct way to stabilize price and frequency. \\
E. Reduce both thermal generation and battery output, because lower controllable supply will automatically reduce demand-side pressure.

\textbf{Output:} Please output the correct candidate answer.
\end{tcolorbox}
\caption{Example of the Acting task.}
\label{fig:prompt_template_acting}
\end{figure}

\clearpage
\end{document}